\documentclass{article}

\usepackage{arxiv}
\usepackage[utf8]{inputenc} 
\usepackage[T1]{fontenc}    
\usepackage{hyperref}       
\usepackage{url}            
\usepackage{booktabs}       
\usepackage{amsfonts}       
\usepackage{nicefrac}       
\usepackage{microtype}      
\usepackage{lipsum}		    
\usepackage{natbib}
\usepackage{doi}

\usepackage{amssymb}
\usepackage{gensymb}
\usepackage{multirow}
\usepackage{color,array,dcolumn}
\usepackage{fixltx2e}
\usepackage{stfloats}
\usepackage{url}
\usepackage{amssymb}
\usepackage{gensymb} 
\usepackage{multirow} 
\usepackage{algorithm, algpseudocode}%

\usepackage{enumitem}
\usepackage{amsmath}
\usepackage{graphicx}
\usepackage{multirow,bigdelim}
\usepackage{booktabs}
\usepackage{textcomp}
\usepackage{adjustbox}
\usepackage{verbatim}
\usepackage{enumitem}
\usepackage{caption}
\usepackage{subcaption}
\usepackage{mathtools}
\usepackage{tikz}
\usetikzlibrary{graphs, graphs.standard, quotes}
\usepackage{todonotes}
\usepackage{wrapfig}
\usepackage{color}
\usepackage[normalem]{ulem}

\usepackage{makecell}

\newcommand{\vect}[1]{\boldsymbol{\mathrm{#1}}}
\algnewcommand{\LineComment}[1]{\State \(\triangleright\) #1}

\floatname{algorithm}{Algorithm}

\title{Online Calibration of Deep Learning Sub-Models for Hybrid Numerical Modeling Systems}

\author{%
  Said Ouala$^{1}$, Bertrand Chapron$^2$, \textbf{Fabrice Collard}$^3$, \textbf{Lucile Gaultier}$^3$, \textbf{Ronan Fablet}$^1$\\
   $(1)$ IMT Atlantique; Lab-STICC, 29200 Brest, France\\
   $(2)$ Ifremer, LOPS, 29200 Brest, France\\
   $(3)$ ODL, 29200 Brest, France\\
  \texttt{\{said.ouala, ronan.fablet\}@imt-atlantique.fr}\\
  \texttt{Bertrand.Chapron@ifremer.fr},\\
  \texttt{\{dr.fab, lucile.gaultier\}@oceandatalab.com}
}

\renewcommand{\shorttitle}{Online Calibration of Deep Learning Sub-Models for Hybrid Numerical Modeling Systems}

\hypersetup{
pdftitle={Online Calibration of Deep Learning Sub-Models for Hybrid Numerical Modeling Systems},
pdfsubject={math.NA, stat.ML},
pdfauthor={Said Ouala, Bertrand Chapron, Ronan Fablet},
pdfkeywords={Hybrid models, Online learning, Numerical models, Parameterization},
}

\usepackage{amsthm}

\newtheorem{proposition}{Proposition}[section]
\newtheorem{theorem}{Theorem}[section]
\newtheorem{corollary}{Corollary}[theorem]

\theoremstyle{example}
\newtheorem{example}{Example}[section]

\begin{document}
\maketitle

\begin{abstract}
Artificial intelligence and deep learning are currently reshaping numerical simulation frameworks by introducing new modeling capabilities. These frameworks are extensively investigated in the context of model correction and parameterization where they demonstrate great potential and often outperform traditional physical models. Most of these efforts in defining hybrid dynamical systems follow {offline} learning strategies in which the neural parameterization (called here sub-model) is trained to output an ideal correction. Yet, these hybrid models can face hard limitations when defining what should be a relevant sub-model response that would translate into a good forecasting performance. End-to-end learning schemes, also referred to as online learning, could address such a shortcoming by allowing the deep learning sub-models to train on historical data. However, defining end-to-end training schemes for the calibration of neural sub-models in hybrid systems requires working with an optimization problem that involves the solver of the physical equations. Online learning methodologies thus require the numerical model to be differentiable, which is not the case for most modeling systems. To overcome this difficulty and bypass the differentiability challenge of physical models, we present an efficient and practical online learning approach for hybrid systems. The method, called EGA for Euler Gradient Approximation, assumes an additive neural correction to the physical model, and an explicit Euler approximation of the gradients. We demonstrate that the EGA converges to the exact gradients in the limit of infinitely small time steps. Numerical experiments are performed on various case studies, including prototypical ocean-atmosphere dynamics. Results show significant improvements over offline learning, highlighting the potential of end-to-end online learning for hybrid modeling.
\end{abstract}

\section{Introduction}
\label{sec:introd}

High-fidelity simulations of physical phenomena require intense modeling and computing efforts. Prohibitive computational costs often lead to only resolving certain spatiotemporal scales and the resulting scale truncation may question the accuracy and reliability of the simulations. For example, in ocean-atmosphere and climate systems, many physical, biological, and chemical phenomena happen at scales finer than the discretization of numerical models. These unresolved processes generally encompass dissipation effects, but are also responsible for some energy redistribution and backscattering \cite{frederiksen1997eddy,shutts2005kinetic,juricke2020ocean}. In practice, parameterizations or sub-models of these phenomena are coupled to the numerical models. These sub-models introduce a significant source of uncertainty and might limit the predictability of large-scale models.

In recent years, scientific computing problems have been explored from the perspective of machine learning and Artificial Intelligence (AI). The combination of AI with computational sciences has given rise to a wide spectrum of methodological questions, articulated in the framework of Scientific Machine Learning (SciML). These questions encompass various aspects, from model acceleration through AI solvers \cite{raissi2019physics, ouala2021learning,partee2022using,kovachki2023neural}, to the learning of model correction terms or sub-models in numerical simulations \cite{maulik2019sub,zanna2020data,gupta2021neural,charalampopoulos2022machine,bonnet2022airfrans,subel2023explaining}. Additionally, Scientific Machine Learning has significantly advanced the development of state-of-the-art surrogate modeling techniques \cite{wang2018model,brunton2019data,ouala2020learning, hasegawa2020machine, ouala2023extending, serrano2023infinity}. These models show particular interest for complex systems where physical models do not exist, or for the replacement of complex representations of physical systems (e.g., computationally expensive computer models) with simpler, computationally-efficient counterparts, capable of capturing some of the underlying dynamics.

Surrogate modeling finds applications in forecasting \cite{ouala_sea_2018,cheng2022data,migus2023stability,serrano2023infinity}, simulation \cite{nguyen2019emlike,brajard2019combining,bocquet2020bayesian,ouala2020learning,ouala2023bounded}, data assimilation \cite{Ouala_2018,cheng2023machine,ouala2023end,cheng2024efficient} and control \cite{kaiser2018sparsity,brunton2021machine}. In the context of Earth system predictability, AI has exhibited remarkable potential and has rapidly gained momentum in leveraging atmosphere-reanalyzed data to provide efficient forecasting systems that are purely data-driven. By learning from relatively coarse‐grained, but fully consistent, space-time atmosphere parameter fields, AI techniques show large promises and often outperform advanced numerical weather forecast models when focusing on specific processes or variables \cite{pathak2022fourcastnet,lam2022graphcast,dueben2022challenges,chen2023fengwu, bi2023accurate,nguyen2023climax}. 

To scale up to entire systems and coupled systems, the design of hybrid modeling approaches is also actively under development, building on the combination of physical equations and deep learning components \cite{jiang2020improving,camps2020advancing,brajard2021combining,guan2022stable,guan2023learning,jakhar2023learning,farchi2023online,shen2023differentiable}. Such strategies require two main aspects. The availability of a \textit{representative enough} database and the ability to translate the problem into a relevant objective function that can be optimized with respect to the deep learning model. From a machine learning point of view, the ability to define end-to-end learning strategies is a key element that allowed deep learning models to advance several signal processing fields. Specifically, questions such as how to train end-to-end a large language model and how to define an objective function on the \textit{usually intractable} likelihood of the data in generative modeling is one of the most important ingredients in the design and success of deep learning architectures. Recent studies, referred to here as online learning schemes, have explored such a philosophy in the calibration of hybrid numerical models that contain both physical and deep learning components 
\cite{um2020solver,frezat2022posteriori,frezat2023gradientfree}. This learning strategy requires, in general, having access to the gradient of the physical model with respect to the parameters of the deep learning component. However, this gradient is usually unavailable for most state-of-that numerical models, which explains why {offline} learning strategies have been mostly considered in hybrid numerical modeling systems \cite{frezat2020physical,zanna2020data,ross2023benchmarking,guan2023learning}.

In this work, we address the online learning of hybrid models. We derive an easy-to-use workflow to perform online optimization while bypassing the differentiability bottleneck of physical models. Our methodology relies on the additive decomposition of the neural sub-model and on an explicit Euler approximation of the gradient. We prove that the proposed Euler Gradient Approximation (EGA) converges to the exact gradients as the time step tends to zero. We validate our contribution through several numerical experiments, including ocean-atmosphere prototypical eddy-resolved simulations. Overall, we show significant improvements compared to standard offline learning methods, achieving performance similar to solving the exact online learning problem. We also show that the proposed methodology can be used to improve sub-models that are trained offline, in producing more realistic simulations. 

The paper is organized as follows. We start in section \ref{sec:prob_form} by giving a general introduction to the parameterization problem in earth system modeling and we discuss and situate some state-of-the-art approaches with respect to deep learning models. We then introduce both offline and online learning of sub-models and present some state-of-the-art solutions to solve the online learning problem including the use of differentiable emulators \cite{nonnenmacher2021deep}, and the definition of appropriate adjoints in {\em optimize than discretize} learning strategies \cite{chen2018neural}. Section \ref{sec:EuApprox} presents the proposed framework, followed by the experiments and results in Section \ref{sec:Experiments}. We discuss perspectives for future works in Section \ref{sec:scoop_lims} and close the paper with a conclusion in Section \ref{seq:conc}.

\begin{figure*}
\centering
\includegraphics[width=\textwidth]{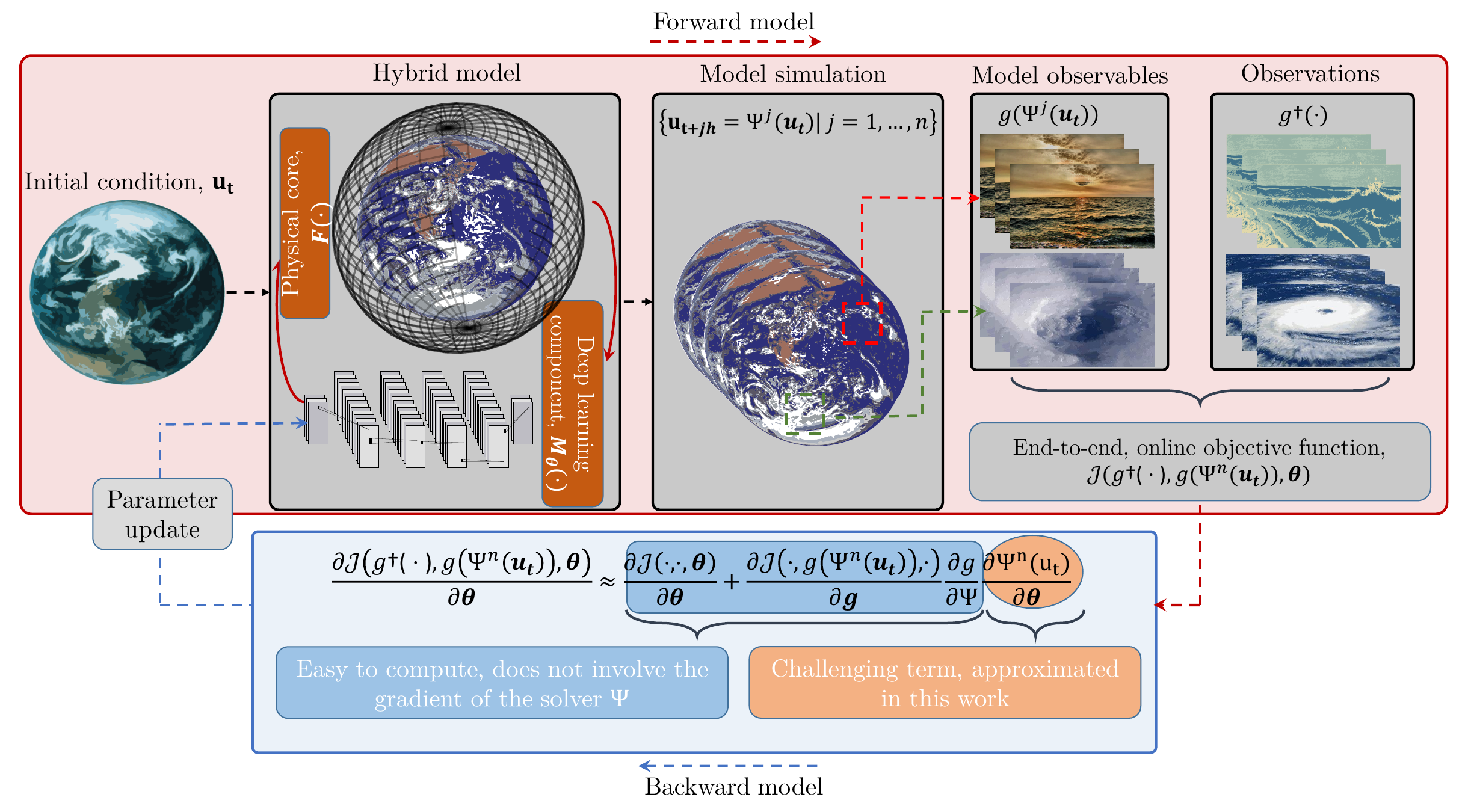}\label{fig:high_fig}
\caption{{{{ \bf  \em Global overview of the online learning problem and link to the proposed EGA.} The aim of the proposed framework is to provide a simple and easy-to-use workflow to solve online optimization problems of hybrid models that contain both a physical core and a deep learning component.}
}}
\label{fig:highlight}
\end{figure*}

\section{Problem formulation and related works}
\label{sec:prob_form}
Hereafter, the physical system of consideration is governed by a differential equation of the following form: 
\begin{equation}
      \partial_t {u}^{\dagger}(t,\cdot) = \mathcal{N}({u}^{\dagger}(t,\cdot)) + \mathcal{G}(\cdot)
  \label{eq:PDE_True}
\end{equation}
where the state ${u}_t^{\dagger} \triangleq {u}^{\dagger}(t,\cdot) \in \mathcal{U}^{\dagger}$ belongs to an appropriate function space $\mathcal{U}^{\dagger}$ for all times $t \geq 0$. The operator $\mathcal{N}$ represents the natural variability of the system and $\mathcal{G}$ is an external forcing. 

This equation describes how a real physical system changes over time. For complex systems such as the atmosphere and the ocean, such models are not available. As an alternative, we must rely on a physical core $\mathcal{F}$, that is coupled with a number of heuristic process representations, encoded in a sub-model $\mathcal{M}_{\vect{\theta}}$ \cite{hourdin2017art}. This sub-model $\mathcal{M}_{\vect{\theta}}$ depends on a vector of parameters $\vect{\theta} \in \mathbb{R}^a$ that need to be tuned using observables of the true state ${u}^{\dagger}$. Both $\mathcal{F}$ and $\mathcal{M}_{\vect{\theta}}$ are (potentially) non-linear operators, constrained to produce well-posed solutions of the state ${u}_t \triangleq {u}(t,\vect{x}) \in \mathcal{U}$ in some appropriate function space $\mathcal{U}$. The time evolution of this state is described by the following Partial Differential Equation (PDE):
\begin{equation}
  \left\{
    \begin{array}{l l}
      \partial_t {u}(t,\vect{x}) &= \mathcal{F}({u}(t,\vect{x})) + \mathcal{M}_{\vect{\theta}}({u}(t,\vect{x}))\\
      {u}(0,\vect{x}) &= {u}_0(\vect{x})
    \end{array}
  \right.
  \label{eq:PDE_Init}
\end{equation}
where $\vect{x} \in \Omega \subset \mathbb{R}^d$, $d \in \mathbb{N}$ is the space variable. The PDE \eqref{eq:PDE_Init} should also be
supplied with appropriate boundary conditions on $\partial \Omega$. 

Defining and calibrating sub-models is a crucial step in physical simulations. Physics-based sub-models rely on first principles to describe the operator $\mathcal{M}_{\vect{\theta}}$. Typical examples can be found in the context of subgrid-scale representations that are based on eddy viscosity assumptions. In such representations, missing scales of motion are assumed to be mainly diffusive and the operator $\mathcal{M}_{\vect{\theta}}$ becomes a diffusion operator \cite{smagorinsky1963general,germano1991dynamic,moin1991dynamic,spalart1992one,wilcox2008formulation}. In the same context, stochastic subgrid scale sub-models, like those developed under the Location Uncertainty \cite{memin2014fluid} (LU) or Stochastic Advection by Lie Transport (SALT) \cite{holm2015variational} paradigms, also rely on physical principles. These sub-models encode within $\mathcal{M}_{\vect{\theta}}$ a diffusive effect of large-scale components by small-scale velocity, small-scale energy backscattering, and a modified advection term related to the turbophoresis phenomenon \cite{resseguier2017geophysicala, resseguier2017geophysicalb, resseguier2017geophysicalc}.

Physics-based sub-models have the advantage of being expressed in a continuous form, enabling theoretical validation and ensuring the equations are well-posed under these sub-models. In contrast, machine learning solutions often rely on discrete versions of the system \eqref{eq:PDE_Init}, which can pose challenges for theoretical validation. Despite recent progress in learning neural operators \cite{kovachki2023neural, lanthaler2023curse}, validating these models remains more complex when compared to physics-based approaches. However, empirical evidence continuously demonstrates the interest in using machine learning models to enhance current state-of-the-art physical simulations.

\subsection{Deep learning and sub-model calibration}

For most standard deep learning models, the models and calibration techniques are built after discretizing the governing equations \eqref{eq:PDE_Init}:
\begin{equation}
  \left\{
    \begin{array}{l l}
      \Dot{\vect{u}}_t &= \vect{F}(\vect{u}_t) + \vect{M}_{\vect{\theta}}(\vect{u}_t)\\
      \vect{u}(0) &= \vect{u}_0
    \end{array}
  \right.
  \label{eq:PDE_disc}
\end{equation}
where $\vect{u}_t \triangleq {\vect{u}}(t) \in L \subset \mathbb{R}^{d_{\vect{u}}}$ is a discretized version of $u_t$ and $\vect{F}$ the corresponding vector field. The sub-model $\vect{M}_{\vect{\theta}}$ is a deep neural network with parameters $\vect{\theta} \in \mathbb{R}^{a}$. The boundary conditions of the problem are dropped for simplicity. From this equation, we also define a flow:
\begin{equation}
      \vect{u}_{t_f} = \phi_{t_f}(\vect{u}_{t}) = \vect{u}_{t} + \int_{t}^{t_f} (\vect{F} + \vect{M}_{\vect{\theta}}) (\vect{u}_w) dw
  \label{eq:flow_mdl}
\end{equation}
Let us also assume that the numerical integration of \eqref{eq:flow_mdl} is performed using a numerical scheme $\Psi$ that runs for the sake of simplicity with a fixed step size $h$:
\begin{equation}
      \vect{u}_{t + n h} = \Psi^n(\vect{u}_{t}) = \underbrace{\Psi \circ \Psi \circ \cdots \circ  \Psi }_{n\text{ times}}(\vect{u}_{t})
  \label{eq:flow_discretized_equation}
\end{equation}
where $n$ is the number of grid points such as $t_f = t + nh$. We assume that the solver $\Psi$ as well as the time step $h$ are defined by some stability and performance criteria that make the resolution of the equation \eqref{eq:flow_discretized_equation} convergent to the true solution \eqref{eq:flow_mdl}.

In this work, we discuss possible optimization strategies for the calibration of $\vect{\theta}$. The sub-model $\vect{M}_{\vect{\theta}}$ is assumed to be a deep learning model, and we focus on gradient-based optimization strategies for calibrating its parameters. Note that calibrating the parameters of $\vect{M}_{\vect{\theta}}$ using gradient-free techniques is also possible. For instance by defining a state space model on the parameters $\vect{\theta}$ and running an ensemble Kalman inversion \cite{iglesias2013ensemble,kovachki2019ensemble,bottcher2023gradient}. However, we focus on gradient based techniques, as they are the most used optimization techniques for calibrating deep learning models. We also assume that there is a single sub-model $\vect{M}_{\vect{\theta}}$, and that its impact on $\vect{F}$ is additive. These assumptions are realistic in several simulation based scenarios and we discuss in section \ref{sec:scoop_lims}, how they can be relaxed.

\subsection{Offline learning strategy}
\label{seq:offline_learning}
The common strategy for optimizing deep learning-based sub-models is to formulate the optimization problem as a supervised regression problem. Provided a reference $\vect{R}_{t} \in \mathbb{R}^{d_{\vect{u}}}$ that represents an ideal sub-model response to the input $\vect{u}_{t}$, the offline learning strategy can be written as the emulation of $\vect{R}_{t}$ using the deep learning model $\vect{M}_{\vect{\theta}}$. This translates into a supervised regression problem where the objective function can be written as:
\begin{equation}
\label{eq:offline_learning}
   \mathcal{Q}(\vect{R}_{t}, \vect{M}_{\vect{\theta}},\vect{\theta}) 
\end{equation}
with $\mathcal{Q}$, a cost function that depends implicitly on $\vect{\theta}$ through $\vect{M}_{\vect{\theta}}$, and possibly, explicitly when, for instance, regularization constraints on the weights of the sub-models are used. The parameters are typically optimized using a stochastic gradient descent algorithm, and the gradients of the model are computed using automatic differentiation.

\begin{example}[Offline learning of subgrid scale sub-models] In subgrid scale modeling, the sub-model $\vect{M}_{\vect{\theta}}$ accounts for unresolved processes due to limited resolution when discretizing the equations \cite{bose2023accurate,guan2023learning,ross2023benchmarking}. In this context, $\vect{R}_t$ is defined as the difference between a high-resolution equation and a low-resolution one i.e., assuming some reference discretization at $\mathbb{R}^{d_{\vect{u}^{\dagger}}}$ with $d_{\vect{u}^{\dagger}} >> d_{\vect{u}}$ in which the continuous equations are perfectly resolved, $\vect{R}_t$ is the subgrid-scale term that is defined as follows 
\begin{equation}
\label{eq:subgrid_scale_R}
   \vect{R}_t = \tau(\vect{F}_{d_{\vect{u}^{\dagger}}}(\vect{u}^{\dagger}_t)) - \vect{F}(\tau({\vect{u}^{\dagger}_t}))
\end{equation}
where $\vect{u}^{\dagger}$ is the true, high resolution, state, $\vect{F}_{d_{\vect{u}^{\dagger}}}$ the equations defined on the high-resolution grid and $\tau$ is a projection from the high-resolution space to the low resolution one. This operation typically includes a filtering of the finer scales in  $\mathbb{R}^{d_{\vect{u}^{\dagger}}}$ and a coarse-graining to the grid in $\mathbb{R}^{d_{\vect{u}}}$. Given a dataset of $N$ snapshot pairs $\{(\vect{R}_{t_k}, \tau({\vect{u}^{\dagger}_{t_k}})) | k = 1, \cdots, N\}$, the cost function can be written as:
\begin{equation}
\label{eq:offline_learning_exple}
   \mathcal{Q}(\vect{R}_{t}, \vect{M}_{\vect{\theta}},\vect{\theta})  = \frac{1}{N}\sum_k \left \|  \vect{R}_{t_k}  -  \vect{M}_{\vect{\theta}}(\tau({\vect{u}^{\dagger}_{t_k}}))   \right \|^2 + \mathcal{R}(\cdot)
\end{equation}
where $\mathcal{R}(\cdot)$ is a regularization term and $\| \cdot \|$ is some appropriate norm (typically $L^2$) in $\mathbb{R}^{d_{\vect{u}}}$. The optimization problem \eqref{eq:offline_learning_exple} is typically solved using a stochastic gradient descent algorithm.    
\end{example}

The offline optimization problem aims to minimize the discrepancy between the sub-model $\vect{M}_{\vect{\theta}}$ and a reference dataset that is considered an ideal response. However, several works showed that, while deep learning models show great success in achieving better offline results than state-of-the-art physical parameterizations, these models might perform poorly in online tests, in which the data-driven sub-model is coupled to the numerical solver of the physical model. These models can show pathological behaviors such as numerical instabilities or physically unrealistic flows that can lead to a blow-up of the simulation \cite{jakhar2023learning,chattopadhyay2023long}. Discussing and accessing the performance of offline closures in online tests is mandatory for the progress and reliability of such tools, and several works have shown that offline sub-models can be constrained to deliver stable long-term online simulations. For instance, \cite{guan2022stable} showed that offline subgrid scale sub-models can produce realistic and stable simulations by using large datasets, and \cite{frezat2020physical,guan2023learning} studied the positive impact of including physical constraints on offline closures to allow for realistic long time simulations. 

Another problem with offline optimization is that for several applications that involve making numerical models closer to historical observations, such as weather forecasting, for instance, the complexity of the underlying dynamics and our lack of prior knowledge makes challenging the definition and construction of an offline reference dataset $\vect{R}_{t_k}$. In such scenarios, relying on end-to-end learning is appealing and has already motivated several works that rely on pure data-driven surrogate models \cite{pathak2022fourcastnet,lam2022graphcast,dueben2022challenges,chen2023fengwu, bi2023accurate,nguyen2023climax}. In the context of hybrid models, a major drawback of such optimization strategies is the inclusion of the numerical solver within the optimization problem, which makes the resolution more challenging. In the rest of this paper, we briefly discuss various online optimization strategies and propose an easy-to-use workflow for learning online sub-models.

\subsection{Online learning strategy:} 

In online learning strategies, the sub-model must minimize the cost between a numerical integration of the model \eqref{eq:flow_discretized_equation} and some reference. Formally, let us assume $\vect{g}^{\dagger} \in \mathcal{F} : \mathcal{U}^{\dagger} \longrightarrow \mathbb{R}^m$ to be a real vector valued observable of the true dynamics described in \eqref{eq:PDE_True}. This observable might correspond to real observations of a geophysical state or to some observable of an idealized experiment for which \eqref{eq:PDE_True} is assumed to be known. To this observable, we associate a real vector valued observable of the dynamical system \eqref{eq:PDE_disc} $\vect{g} \in \mathcal{F} : L \longrightarrow \mathbb{R}^m$. The online learning problem aims at matching these two quantities using an objective function of the following form:
\begin{equation}
\mathcal{J}(\vect{g}^{\dagger}({u}^{\dagger}_{t+nh}), \vect{g}(\Psi^n(\vect{u}_{t})),\vect{\theta})\label{eq:online_loss}
\end{equation}
where $n$ is a hyperparameter that corresponds to the number of integration time steps of \eqref{eq:flow_discretized_equation} used in the optimization. Here, the cost function $\mathcal{J}$ is given in discrete time and depends on the solution of the dynamical model \eqref{eq:flow_discretized_equation}.

The online learning strategy significantly widens the range of metrics that can be considered to calibrate the sub-model parameters. It can indeed link the training of the sub-model parameters to any end-to-end constraint related to the (short-term) simulation of the dynamical model \eqref{eq:PDE_disc}.

\begin{example}[Online learning of subgrid scale sub-models] Similarly to the previous example, the sub-model $\vect{M}_{\vect{\theta}}$ is assumed to account for unresolved processes due to limited resolution after discretizing the equations. In this context, ${u}^{\dagger} \in \mathbb{R}^{d_h}$ is the true reference high-resolution simulation, and $\vect{g}^{\dagger}$ is some observable on this high-resolution simulation. A natural choice of this observable, e.g. \cite{frezat2022posteriori}, is a filtering and coarse graining operation, {\em i.e.} $\vect{g}^{\dagger} = \tau$. The observable on the low-resolution model $\vect{g}$ is a full state observable, {\em i.e.,} $\vect{u} = \vect{g}(\vect{u})$, and the online loss function can be simplified and defined as the difference between a high-resolution equation and the low resolution one. Given a dataset of $N$ pairs of time series $\{(\tau(u^{\dagger}_{t_k +jh}), \tau(u^{\dagger}_{t_k}))| \text{ with } k = 1 \dots N \text{ and } j = 1 \dots n\}$, the cost function can write: 
\begin{equation}
\label{eq:subgrid_scale_online}
\mathcal{J}(\cdot) =  \frac{1}{N}\sum_k \frac{1}{n}\sum_{j = 1}^{n} \| \tau({u}^{\dagger}_{t_{k} + jh}) - \Psi^j(\tau(u^{\dagger}_{t_k})) \| + \mathcal{R}(\cdot) 
\end{equation}   
\end{example}

The resolution of the optimization problem \eqref{eq:online_loss} becomes challenging, as it now involves the resolution of the whole dynamical system \eqref{eq:flow_mdl}. Specifically, when solving the online optimization problem, i.e., $\arg \min_{\vect{\theta}} \mathcal{J}$, using gradient descent, it is necessary to compute the gradient of the loss \eqref{eq:online_loss} with respect to the parameters of the sub-model:
\begin{equation}
\label{eq:grad_online_true}
\begin{aligned}
    \frac{\partial \mathcal{J}}{\partial\vect{\theta}}(\vect{g}^{\dagger}({u}^{\dagger}_{t+nh}), \vect{g}(\Psi^n(\vect{u}_{t})),\vect{\theta}) & = \frac{\partial \mathcal{J}(\cdot,\cdot,\theta)}{\partial\vect{\theta}} + \frac{\partial \mathcal{J}(\cdot,\vect{g}(\Psi^n(\vect{u}_{t})),\cdot)}{\partial \vect{g}}  \frac{\partial \vect{g}}{\partial \Psi}\frac{\partial \Psi^n(\vect{u}_{t})}{\partial\vect{\theta}}
\end{aligned}
\end{equation}
The gradient of the solver $\Psi^n$ must thus be evaluated for every $n$ with respect to the parameters of the sub-model: 
\begin{equation}
\frac{\partial \Psi^n(\vect{u}_{t})}{\partial\vect{\theta}} = \frac{\partial}{\partial\vect{\theta}} \Psi \circ \Psi \circ \cdots \circ  \Psi(\vect{u}_{t})
\label{eq:grad_of_phi}
\end{equation}

Computing this gradient requires the solver in $\Psi$ to be differentiated with respect to the parameters of the sub-model, which critically limits the use of the online approach in practice. Most large-scale forward solvers in Earth system models (ESM), and digital twin frameworks in Computational Fluid Dynamics (CFD) applications rely on high-performance languages and tools that do not embed automatic differentiation (AD). They can not provide adjoint models that evaluate gradients with respect to deep learning parameters. This largely explains the extensive emphasis on offline calibration schemes, even though, the aim of a sub-model is always to have a good online performance.
\subsection{Emulators}
\label{sec:emulators}

In our context, differentiable emulators \cite{nonnenmacher2021deep} are considered to both approximate the physical part of \eqref{eq:PDE_disc} and the numerical solver $\Psi$. These models were successfully tested on data assimilation toy problems \cite{nonnenmacher2021deep}, and can be adapted to the online learning of sub-models \cite{frezat2023gradientfree}. 

Let $\vect{G}_{\vect{\phi}}$ be a deep learning approximation of $\vect{F}$ such that $\Dot{\vect{u}}_t \approx   \vect{G}_{\vect{\phi}}(\vect{u}_t) + \vect{M}_{\vect{\theta}}$ and let $\Psi_{\vect{\alpha}}$ be the numerical scheme with parameters $\vect{\alpha}$ that solves this approximate model. The solver $\Psi_{\vect{\alpha}}$ can be based on a deep learning model. For the sake of simplicity of presentation, the solver $\Psi_{\vect{\alpha}}$ is assumed to be an explicit single-step scheme, discretized with a fixed step size $h'$. The emulator aims at approximating the true solver \eqref{eq:flow_discretized_equation} {\em i.e.}:
\begin{equation}
\begin{aligned}
\label{eq:emulators}
\vect{u}_{t_f} =  \vect{u}_{t_0 + n h} &= \underbrace{\Psi \circ \Psi \circ \cdots \circ  \Psi }_{n\text{ times}}(\vect{u}_{t_0})\\
&\approx \underbrace{\Psi_{\vect{\alpha}} \circ \Psi_{\vect{\alpha}} \circ \cdots \circ  \Psi_{\vect{\alpha}} }_{r\text{ times}}(\vect{u}_{t_0})
\end{aligned}
\end{equation}

Assuming that $r$, $h'$, $\vect{G}_\phi$ and $\Psi_{\vect{a}}$ are correctly calibrated, the gradients of the true solution with respect to the parameters are replaced by the ones of \eqref{eq:emulators}. In practice, $\vect{G}_\phi$ (and occasionally $\Psi_{\vect{\alpha}}$) as well as the sub-model $\vect{M}_{\vect{\theta}}$ can be trained sequentially. 
For complex models, these emulators, and their solvers, require careful tuning to ensure that the training of the sub-model is not biased by the uncertainty of the emulator \cite{frezat2023gradientfree}.

\subsection{Link to optimal control and gradient descent by the shooting method}
The online optimization problem can be written as an optimal control problem with the following optimization problem:
\begin{equation}
  \left\{
    \begin{array}{c l}
                    &\arg \min_{\vect{\theta}} \mathcal{J}\\
      \text{s.t. }  &\Dot{\vect{u}}_t = \vect{F}(\vect{u}_t) + \vect{M}_{\vect{\theta}}(\vect{u}_t)
    \end{array}
  \right.
  \label{eq:opti_cont}
\end{equation}

A Lagrangian of this optimization problem can then be written as: 
\begin{equation}
\mathcal{L} = \mathcal{J} + \int_{t}^{t+nh} \lambda_t (\Dot{\vect{u}}_t -  \vect{F}(\vect{u}_t) - \vect{M}_{\vect{\theta}}(\vect{u}_t)) dt
\label{eq:lagrangian}
\end{equation}
where $\lambda_t$ is a time-dependent Lagrange multiplier. 
The gradients of the online objective function with respect to the parameters are identical to the ones of the Lagrangian. The optimal control problem is here formulated in continuous time, but a corresponding discrete time control can also be defined based on the discretized dynamics \eqref{eq:flow_discretized_equation}. 

Interpreting this machine learning problem as an optimal control provides multiple methods to build adjoint models of \eqref{eq:flow_discretized_equation} with respect to the parameters of the deep learning sub-model. Overall, optimal control methods can be divided into direct and indirect approaches \cite{andersson2013general,andersson2019casadi}. Direct approaches, {\em discretize then optimize} methods, are based on a discrete dynamical model constraint. Indirect approaches, {\em optimize then discretize} techniques, work on the continuous optimal control problem \eqref{eq:opti_cont}. The latter technique was advertised in \cite{chen2018neural} as a (potentially) efficient methodology for the optimization of neural ODE models. 

In practice, one significant drawback of employing optimal control methods lies in the complexity of designing the adjoint model. It necessitates a domain-specific treatment based on the underlying dynamics of the physical equations, which might limit the use of this strategy in the context of online learning of deep learning sub-models.

\section{Euler Gradient Approximation (EGA) for the online learning problem}
\label{sec:EuApprox}

We aim to define a relevant, easy-to-use workflow for solving the online optimization problem in hybrid modeling systems where the physical model is not differentiable. In order to do so, we introduce the EGA, which relies on an explicit Euler discretization for the computation of the gradients, {\em i.e.} the backward pass in deep learning. 

Recall that the online cost function \eqref{eq:online_loss}:
\begin{equation*}
\mathcal{J}(\vect{g}^{\dagger}({u}^{\dagger}_{t+nh}), \vect{g}(\Psi^n(\vect{u}_{t})),\vect{\theta})
\end{equation*}
depends explicitly on the non-differentiable solver $\Psi$, that runs (for the sake of simplicity) with a fixed step size $h$:
\begin{equation*}
      \vect{u}_{t_f} =  \vect{u}_{t + nh} = \Psi^n(\vect{u}_{t}) =  \underbrace{\Psi \circ \Psi \circ \cdots \circ  \Psi }_{n\text{ times}}(\vect{u}_{t})
\end{equation*}
As discussed in the previous section, the resolution of the online optimization problem requires the computation of the gradient of the online objective with respect to the parameters of the sub-model. By using the chain rule, it is trivial to show that this gradient depends on the gradient of the solver $\Psi$ {\em i.e.}:
\begin{equation*}
\begin{aligned}
    \frac{\partial \mathcal{J}}{\partial\vect{\theta}}(\vect{g}^{\dagger}({u}^{\dagger}_{t+nh}), \vect{g}(\Psi^n(\vect{u}_{t})),\vect{\theta}) & = \frac{\partial \mathcal{J}(\cdot,\cdot,\theta)}{\partial\vect{\theta}} + \frac{\partial \mathcal{J}(\cdot,\vect{g}(\Psi^n(\vect{u}_{t})),\cdot)}{\partial \vect{g}}  \frac{\partial \vect{g}}{\partial \Psi}\frac{\partial \Psi^n(\vect{u}_{t})}{\partial\vect{\theta}}
\end{aligned}
\end{equation*}

In order to solve the online optimization problem, we aim to find an efficient approximation of the gradient of the solver $\Psi$ with respect to the parameters of the sub-model. Let us consider an explicit Euler solver $\Psi_E$, a single integration step of \eqref{eq:PDE_disc} using $\Psi_E$ can be written as:
\begin{equation}
\label{eq:psisolverandexpEuler}
\begin{aligned}
    \vect{u}_{t+h} &= \Psi_E(\vect{u}_t)
\end{aligned}
\end{equation}
where $\Psi_E(\vect{u}_{t}) = {\vect{u}}_{t} + h (\vect{F}({\vect{u}}_{t})+\vect{M}_{\vect{\theta}}({\vect{u}}_{t}))$. 

We aim at approximating the gradient of the solver $\Psi$, using the ones of the Euler solver. In this context, and in order to keep track of the order of convergence of the gradient, we introduce the following proposition.
\begin{proposition}
\label{prop:psi_as_euler}
Assuming that the solver $\Psi$ has order $p\geq 1$, we have for any initial condition $\vect{u}_t$:
\begin{equation}
\label{eq:psifuncpsiE}
\begin{aligned}
    \vect{u}_{t+h} &= \Psi(\vect{u}_t) \\
                   &= \Psi_E(\vect{u}_t) + O(h^2)
\end{aligned}
\end{equation}
In particular, every solution at an arbitrary time $t_f = t+nh$ can be written as:
\begin{equation}
\label{eq:psifuncpsiE_compose}
\begin{aligned}
    \vect{u}_{t_f} = \vect{u}_{t+nh} &= \Psi^n(\vect{u}_t) \\
                          &= \Psi_E(\Psi^{n-1}(\vect{u}_t)) + O(h^2)
\end{aligned}
\end{equation}   
\end{proposition}
 
By using the proposition \ref{prop:psi_as_euler}, we can show that the gradient of the solver $\Psi$ can be written as a function of the gradient of the Euler solver plus a bounded term.
\begin{theorem}[EGA]:
\label{theorem:main_result_grad}
Under the same conditions as proposition \ref{prop:psi_as_euler}, and assuming that the number of time steps $n$ is fixed and corresponds to a hyperparameter of the online learning problem, the gradient of the solver $\Psi$ can be written as follows:
\begin{equation}
\label{eq:euler_grad_psitheorem}
\begin{aligned}
\frac{\partial}{\partial\vect{\theta}}  {\Psi^n}(\vect{u}_{t}) &=\sum_{j = 1}^{j = n-1} \Bigl(\prod_{i = 1}^{i = n-j} \underbrace{\frac{\partial \Psi(\Psi^{n-i}(\vect{u}_{t}))}{\partial \Psi^{n-i}(\vect{u}_{t})}}_{\text{Jacobian of the flow}}\Bigl) h \underbrace{\frac{\partial}{\partial\vect{\theta}} \vect{M}_{\vect{\theta}}(\Psi^{j -1}(\vect{u}_{t}))}_{\text{Gradient of the sub-model}} + h\frac{\partial}{\partial\vect{\theta}} \vect{M}_{\vect{\theta}}(\Psi^{n-1}(\vect{u}_{t})) + {O(h^2)}
\end{aligned}
\end{equation}
\end{theorem}

\begin{corollary}
\label{corrol:approx_grad_h}
Under the same conditions as proposition \ref{prop:psi_as_euler}, and if we assume that the online learning problem is defined for a given initial and finite times $t_0$ and $t_f$ such that $n = \frac{t_f-t_0}{h}$, then the convergence of the EGA becomes linear in $h$ {\em i.e.}:
\begin{equation}
\label{eq:gradpsi_corollary}
\begin{aligned}
\frac{\partial}{\partial\vect{\theta}}  {\Psi^n}(\vect{u}_{t}) &= \sum_{j = 1}^{j = n-1} \Bigl(\prod_{i = 1}^{i = n-j} \frac{\partial \Psi(\Psi^{n-i}(\vect{u}_{t}))}{\partial \Psi^{n-i}(\vect{u}_{t})}\Bigl) h \frac{\partial}{\partial\vect{\theta}} \vect{M}_{\vect{\theta}}(\Psi^{j -1}(\vect{u}_{t}))+ h \frac{\partial}{\partial\vect{\theta}} \vect{M}_{\vect{\theta}}(\Psi^{n -1}(\vect{u}_{t}))+O(h)
\end{aligned}
\end{equation}
\end{corollary}

Here, we assume for simplicity that the Euler solver runs at the same time step as the one of the solver $\Psi$. In practice and as shown in the experiments, these solvers might have different time steps and our methodology only requires training trajectories of the solver $\Psi$ to be sampled at the time step of the Euler solver. 

Both theorem \ref{theorem:main_result_grad} and corollary \ref{corrol:approx_grad_h} show that the gradient of the solver $\Psi$ can be approximated by knowing two terms, the Jacobian of the flow as well as the gradient of the sub-model with respect to the parameters $\vect{\theta}$ at some simulation time step of the solver $\vect{u}_{t_k + jh} = \Psi^j(\vect{u}_{t})$ with $j = 1, \dots, n-1$. While the gradient of the sub-model with respect to the parameters can be computed for every input $\vect{u}_{t + jh}$ using automatic differentiation, the Jacobian of the flow needs to be specified or approximated.

\subsection{EGA Jacobian approximation}
The Jacobian in the EGA equations \eqref{eq:euler_grad_psitheorem} and \eqref{eq:gradpsi_corollary} need to be provided in order to evaluate the gradients of the solver $\Psi$ with respect to the parameters. In practice, several techniques can be used, we discuss in this work three techniques based on a static formulation of the Jacobian, on the use of a tangent linear model and on an ensemble approximation.
\subsubsection{Static formulation (Static-EGA)}
A static formulation of the Jacobian corresponds to approximating the Jacobian term in \eqref{eq:euler_grad_psitheorem} and \eqref{eq:gradpsi_corollary} by the identity matrix. It can be shown that a static formulation of the Jacobian matrix implies corollary \ref{corrol:gradpsi_identity_corollary}.
\begin{corollary}[Static-EGA]
\label{corrol:gradpsi_identity_corollary}
Under the same conditions as proposition \ref{prop:psi_as_euler}, and assuming that the number of time steps $n$ is fixed, and corresponds to a hyperparameter of the online learning problem, the gradient of the solver $\Psi$ can be written as the one of the Euler solver as follows:
\begin{equation}
\label{eq:gradpsi_identity_corollary1}
\begin{aligned}
\frac{\partial}{\partial\vect{\theta}}  {\Psi^n}(\vect{u}_{t}) &= \sum_{j = 1}^{j = n} h \frac{\partial}{\partial\vect{\theta}} \vect{M}_{\vect{\theta}}(\Psi^{j -1}(\vect{u}_{t})) + O(h^2)
\end{aligned}
\end{equation} 

If we assume that the number of time steps varies with, $h$ {\em i.e.}, that the online learning problem is defined for a given initial and finite times $t_0$ and $t_f$ such that $n = \frac{t_f-t_0}{h}$ that the error term bound becomes is linear in $h$ {\em i.e.}:
\begin{equation}
\label{eq:gradpsi_identity_corollary2}
\begin{aligned}
\frac{\partial}{\partial\vect{\theta}}  {\Psi^n}(\vect{u}_{t}) &= \sum_{j = 1}^{j = n} h \frac{\partial}{\partial\vect{\theta}} \vect{M}_{\vect{\theta}}(\Psi^{j -1}(\vect{u}_{t})) + O(h)
\end{aligned}
\end{equation} 
\end{corollary}

This static approximation provides a simple, easy-to-use, approximation of the gradient of the solver with respect to the parameters $\vect{\theta}$. Furthermore, and similarly to the gradients in \eqref{eq:euler_grad_psitheorem} and \eqref{eq:gradpsi_corollary}, the order of convergence of the gradients in \eqref{eq:gradpsi_identity_corollary1} and \eqref{eq:gradpsi_identity_corollary2} is quadratic and linear, respectively. However, the constant of convergence of this approximation is larger than the approximations in \eqref{eq:euler_grad_psitheorem} and \eqref{eq:gradpsi_corollary} due to the presence of additional quadratic terms in the $O(h^2)$. In this context, improving the precision of the gradient approximation requires either reducing the time step $h$, or providing a better approximation of the Jacobian matrix.

\subsubsection{Tangent Linear Model (TLM) approach (TLM-EGA)}
The rise of variational data assimilation techniques motivated the development of Tangent Linear Models for several physical models. This TLM corresponds to the Jacobian of the solver $\Psi$ that operates only on the physical core $\vect{F}$, without a sub-model $\vect{M}_{\vect{\theta}}$\footnote{Here, we assume for simplicity that the tangent linear model does not include the variation from any physical sub-model.}. Let us call this solver $\Psi_o$ and it corresponds to the time discretization of the following integral:
\begin{equation}
      \vect{u}_{t_f} = \Psi^n_o(\vect{u}_{t_0}) \approx \vect{u}_{t} + \int_{t}^{t_f} \vect{F} (\vect{u}_w) dw
  \label{eq:flow_mdl_physical}
\end{equation}

The tangent linear model can be defined simply, for every initial condition $\vect{u}_{t}$ as the Jacobian of $\Psi_o$ {\em i.e.}:
\begin{equation}
      TLM(\vect{u}_{t}) = \frac{\partial \Psi_o(\vect{u}_{t})}{\partial \vect{u}_{t}}
  \label{eq:TLM}
\end{equation}

In order to use this TLM in the gradient defined in \eqref{eq:euler_grad_psitheorem} and \eqref{eq:gradpsi_corollary}, we need to add the variation that is due to the sub-model $\vect{M_{\theta}}$. Since the sub-model is additive, we can write the following result.
\begin{corollary}[Extended TLM for Hybrid systems]
\label{corrol:TLM}
Under the same conditions of proposition \ref{prop:psi_as_euler}, and assuming that the order of both solvers $\Psi$ and $\Psi_o$ is $p$, and given the TLM of $\Psi_o$. We can write the Jacobian of $\Psi$ as follows:
\begin{equation}
\label{eq:TLM_corollary_1}
\begin{aligned}
\frac{\partial \Psi(\vect{u}_{t})}{\partial \vect{u}_{t}} &= TLM(\vect{u}_{t}) + \frac{\partial}{\partial \vect{u}_{t}} \sum_{k = 1}^{k = p} \frac{h^k}{k!} \vect{M}_{\vect{\theta}}(\vect{u}_{t})^{(k-1)}+ O(h^{p+1})
\end{aligned}
\end{equation} 
\end{corollary}
Since all the reminder terms in the results showed in \eqref{eq:euler_grad_psitheorem} and \eqref{eq:gradpsi_corollary} are bounded by at most $O(h^2)$, a sufficient approximation of the Jacobian term based on the TLM of $\Psi_o$ requires evaluating the Jacobian of $\vect{M}_{\vect{\theta}}$ only up to $k = 1$ {\em i.e.}:
\begin{equation}
\label{eq:TLM_corollary_2}
\begin{aligned}
\frac{\partial \Psi(\vect{u}_{t})}{\partial \vect{u}_{t}} &= TLM(\vect{u}_{t})  + h \frac{\partial \vect{M}_{\vect{\theta}}(\vect{u}_{t})}{\partial \vect{u}_{t}}+ O(h^{2})
\end{aligned}
\end{equation} 
We recall that the Jacobian of $\vect{M}_{\vect{\theta}}$ can be computed efficiently using automatic differentiation. 

\subsubsection{Ensemble Tangent Linear Model (ETLM) approximation (ETLM-EGA)}
The Jacobian of the flow in both \eqref{eq:euler_grad_psitheorem} and \eqref{eq:gradpsi_corollary} can be approximated by using an ensemble. This formulation is widely employed in data assimilation, and it is documented in numerous state-of-the-art works. For completeness, we provide a brief overview here, which is largely based on the work of \cite{allen2023ensemble}.

For every initial condition $\vect{u}_t$, we start by constructing a $K$-member ensemble {\em i.e.} $\{ \vect{u}^i_t | i = 1 \cdots K\}$. These initial conditions are then propagated by the solver $\Psi$ to compute the ensemble prediction:
\begin{equation}
\label{eq:ensemble_forecast}
\begin{aligned}
\vect{u}^i_{t+h} &= \Psi(\vect{u}^i_{t}) \text{ for $i = 1,\cdots,K$ }
\end{aligned}
\end{equation} 

Perturbations are constructed relative to the ensemble mean (indicated by $\overline{\vect{u}}_t$), $\delta \vect{u}^i_{t} = \vect{u}^i_{t} - \overline{\vect{u}}_{t}$ and $\delta \vect{u}^i_{t + h} = \vect{u}^i_{t + h} - \overline{\vect{u}}_{t + h}$, and these are assembled into $d_{\vect{u}} \times K$ matrices, $\delta \vect{U}_{t}$ and $\delta \vect{U}_{t + h}$, representing ensemble perturbations listed column-wise for times $t$ and $t +h$, respectively. Based on these ensemble perturbations, the Jacobian matrix can be approximated as the best linear fit between $\delta \vect{U}_{t}$ and $\delta \vect{U}_{t + h}$: 
\begin{equation}
\label{eq:EnJacobian}
\begin{aligned}
\frac{\partial \Psi(\vect{u}_{t})}{\partial \vect{u}_{t}} \approx ETLM =  \delta \vect{U}_{t + h} \delta \vect{U}_{t}^T [\delta \vect{U}_{t} \delta \vect{U}_{t}^T]^{-1}
\end{aligned}
\end{equation} 
where $T$ denotes the matrix transpose, and $-1$ represents the matrix inverse. 

The ensemble approximation of the Jacobian is particularly valuable when missing a tangent linear model, as it provides a simple data-driven approximation of the sensitivity of the flow to initial conditions. However, When utilizing an ensemble approximation, a particularly challenging aspect arises when dealing with high-dimensional systems, where the system dimension $d_{\vect{u}}$, is significantly larger than the number of ensemble members $K$. In such situations, a good calibration of the initial perturbation is mandatory in order to produce an informative ensemble. 

\subsection{Implementation in deep learning frameworks}
Based on the above formulations, one can derive approximations for the gradient of the online cost \eqref{eq:online_loss}. This involves substituting the gradient of the solver in \eqref{eq:grad_online_true} with any of the provided methods and neglecting the $O(\cdot)$ term. 
\begin{equation}
\label{eq:grad_online_autodiff}
\begin{aligned}
    \frac{\partial \mathcal{J}}{\partial\vect{\theta}} (\vect{g}^{\dagger}({u}^{\dagger}_{t+nh}), \vect{g}(\Psi^n(\vect{u}_{t})),\vect{\theta}) &= \frac{\partial \mathcal{J}(\cdot,\cdot,\theta)}{\partial\vect{\theta}} + \frac{\partial \mathcal{J}(\cdot,\vect{g}(\Psi^n(\vect{u}_{t})),\cdot)}{\partial \vect{g}}  \frac{\partial \vect{g}}{\partial \Psi}\frac{\partial \Psi^n(\vect{u}_{t})}{\partial\vect{\theta}}\\
    &= \vect{v} + \vect{w}\vect{A}_{l,p} + O(h^p)
\end{aligned}    
\end{equation}
where $\vect{v} = \frac{\partial \mathcal{J}(\cdot,\cdot,\theta)}{\partial\vect{\theta}}$ represents the gradient of the regularization, $\vect{w} = \frac{\partial \mathcal{J}(\cdot,\vect{g}(\Psi^n(\vect{u}_{t})),\cdot)}{\partial \vect{g}}  \frac{\partial \vect{g}}{\partial \Psi}$ is the gradient of the online cost with respect to the solver and $\vect{A}_{l,p}$ is the approximation of the gradient of the solver with respect to the parameters. The subscript $p$ represents the order of approximation of the gradient, it also implicitly specifies the training methodology adopted for the number of time steps $n$. In particular, if $p = 2$, the number of steps $n$ is fixed and is considered as a training hyperparameter. If $p = 1$, it means that $n$ is deduced from a given initial and finite time. The subscript $l$ represents the methodology used to compute the Jacobian of the flow:
\begin{equation*}
\begin{aligned}
    \vect{A}_{l,p} = \sum_{j = 1}^{j = n-1} \vect{J}_{j,l} h \frac{\partial}{\partial\vect{\theta}} \vect{M}_{\vect{\theta}}(\Psi^{j -1}(\vect{u}_{t}))+ h \frac{\partial}{\partial\vect{\theta}} \vect{M}_{\vect{\theta}}(\Psi^{n -1}(\vect{u}_{t})))
\end{aligned}    
\end{equation*}
where $\vect{J}_{j,l} = \Bigl(\prod_{i = 1}^{i = n-j} \frac{\partial \Psi(\Psi^{n-i}(\vect{u}_{t}))}{\partial \Psi^{n-i}(\vect{u}_{t})}\Bigl)$ is defined from a Jacobian approximation as follows:
\begin{equation}
    \frac{\partial \Psi(\Psi^{n-i}(\vect{u}_{t}))}{\partial \Psi^{n-i}(\vect{u}_{t})} = 
\begin{cases}
    \vect{\frac{\partial \Psi(\Psi^{n-i}(\vect{u}_{t}))}{\partial \Psi^{n-i}(\vect{u}_{t})}},& \text{if } l = 1 \text{ (EGA)}\\
    \vect{I},& \text{if } l = 2 \text{ (Static-EGA)}\\
    {TLM},& \text{if } l = 3 \text{ (TLM-EGA, Eqs. \eqref{eq:TLM_corollary_2})}\\
    {ETLM},& \text{if } l = 4 \text{ (ETLM-EGA, Eqs. \eqref{eq:EnJacobian})}\\
\end{cases}
\label{eq:EGA_cases}
\end{equation}

The implementation of \eqref{eq:grad_online_autodiff} in programming languages that support automatic differentiation requires evaluation of the gradient of the sub-model $\vect{M}_{\vect{\theta}}$ for inputs $\Psi^j(\vect{u}_t)$ that are issued from the non-differentiable solver $\Psi$. These gradients are multiplied with the Jacobian of the flow $\vect{J}_{j,l}$ to produce an estimate of the gradient of the solver. The evaluation of the gradient can be based on composable function transforms \cite{bradbury2018jax,functorch2021}, or on the modification of the backward call in standard automatic differentiation languages.  In the appendix (see Appendix \ref{sec:algos}), we provide algorithms that illustrate practical implementations of these two methodologies.

\section{Experiments}
\label{sec:Experiments}
Numerical experiments are performed to evaluate the proposed online learning techniques. We first validate numerically the order of convergence of some results developed in the previous section. We then evaluate sub-models trained online with the proposed gradient approximations. These sub-models are benchmarked against state-of-the-art training techniques, including online learning with the exact gradient (when assuming that the gradient of the solver is available). The comparison of the sub-models is done principally on online metrics, where we evaluate the hybrid system (the physical core coupled to the deep learning sub-model) in simulating trajectories that are realistic with respect to some reference data. We also compare, when relevant, offline metrics, where only the output of the sub-model is evaluated (without considering the coupling to the physical core). We consider two case studies on the Lorenz-63 and quasi-geographic dynamics. Additional experiments on the multiscale Lorenz 96 system \cite{lorenz1996predictability}, as well as the details on the parameterization of the deep learning sub-models, training data, objective functions, and baseline methods are given in the appendix.

\subsection{Lorenz 63 system}
\label{sec:Experiments_L63}
The Lorenz 63 dynamical system is a 3-dimensional model of the form:
\begin{equation}
\label{eq:lorenz-63_vf}
\begin{aligned}
    \Dot{{u}}^{\dagger}_{t,1} &= \sigma({u}^{\dagger}_{t,2}-{u}^{\dagger}_{t,1})\\
    \Dot{{u}}^{\dagger}_{t,2} &= \rho {u}^{\dagger}_{t,1} - {u}^{\dagger}_{t,2} - {u}^{\dagger}_{t,1}{u}^{\dagger}_{t,3}\\
    \Dot{{u}}^{\dagger}_{t,3} &= {u}^{\dagger}_{t,1}{u}^{\dagger}_{t,2} - \beta{u}^{\dagger}_{t,3}
\end{aligned}
\end{equation}
Under parametrization $\sigma =10$, $\rho=28$ and  $\beta=8/3$, this system exhibits chaotic dynamics with a strange attractor \cite{lorenz_deterministic_1963}.

We assume that we are provided with $\vect{F}$, an imperfect version of the Lorenz system \eqref{eq:lorenz-63_vf} that does not include the term $\beta{u}^{\dagger}_{t,3}$. This physical core is supplemented with a sub-model $\vect{M}_{\vect{\theta}}$ as follows:
\begin{equation}
\label{eq:lorenz-63_approx}
\Dot{\vect{u}}_t = \vect{F}(\vect{u}_t) + \vect{M}_{\vect{\theta}}(\vect{u}_t)
\end{equation}
where $\vect{u}_t = [{u}_{t,1},{u}_{t,2},{u}_{t,3}]^T$ and $\vect{F}: \mathbb{R}^3 \longrightarrow \mathbb{R}^3$ is given by:
\begin{equation}
\label{eq:lorenz-63_approx_vf}
\begin{aligned}
    F_1(\vect{u}_t) &= \sigma({u}_{t,2}-{u}_{t,1})\\
    F_2(\vect{u}_t) &= \rho {u}_{t,1} - {u}_{t,2} - {u}_{t,1}{u}_{t,3}\\
    F_3(\vect{u}_t) &= {u}_{t,1}{u}_{t,2}
\end{aligned}
\end{equation}
The sub-model $\vect{M}_{\vect{\theta}}$ is a fully connected neural network with parameters $\vect{\theta}$.

\subsubsection{Analysis of the order of convergence}
\label{exp:order_conv}
Here, we present a numerical validation of the convergence order for both the EGA and Static-EGA results developed in the previous section. We specifically focus on the results of the theorem \ref{theorem:main_result_grad} equation \eqref{eq:euler_grad_psitheorem} and the one of the corollary \ref{corrol:gradpsi_identity_corollary} equation \eqref{eq:gradpsi_identity_corollary1}. In this experiment, we evaluate the error of the proposed gradient approximations when the time step ${h}$ varies from $10^{-1}$ to $10^{-4}$. This time step will correspond to the one used by the Euler approximation of the gradient, and not to the time step of the forward solver $\Psi$. Regarding the solver $\Psi$, we use in this experiment a differentiable adaptive step size solver $\Psi$ (DOPRI8 used in \cite{chen2018neural}). This solver allows us to compute the exact gradient of the online cost function using automatic differentiation.

\begin{figure*}
\centering
\includegraphics[width=0.5\textwidth]{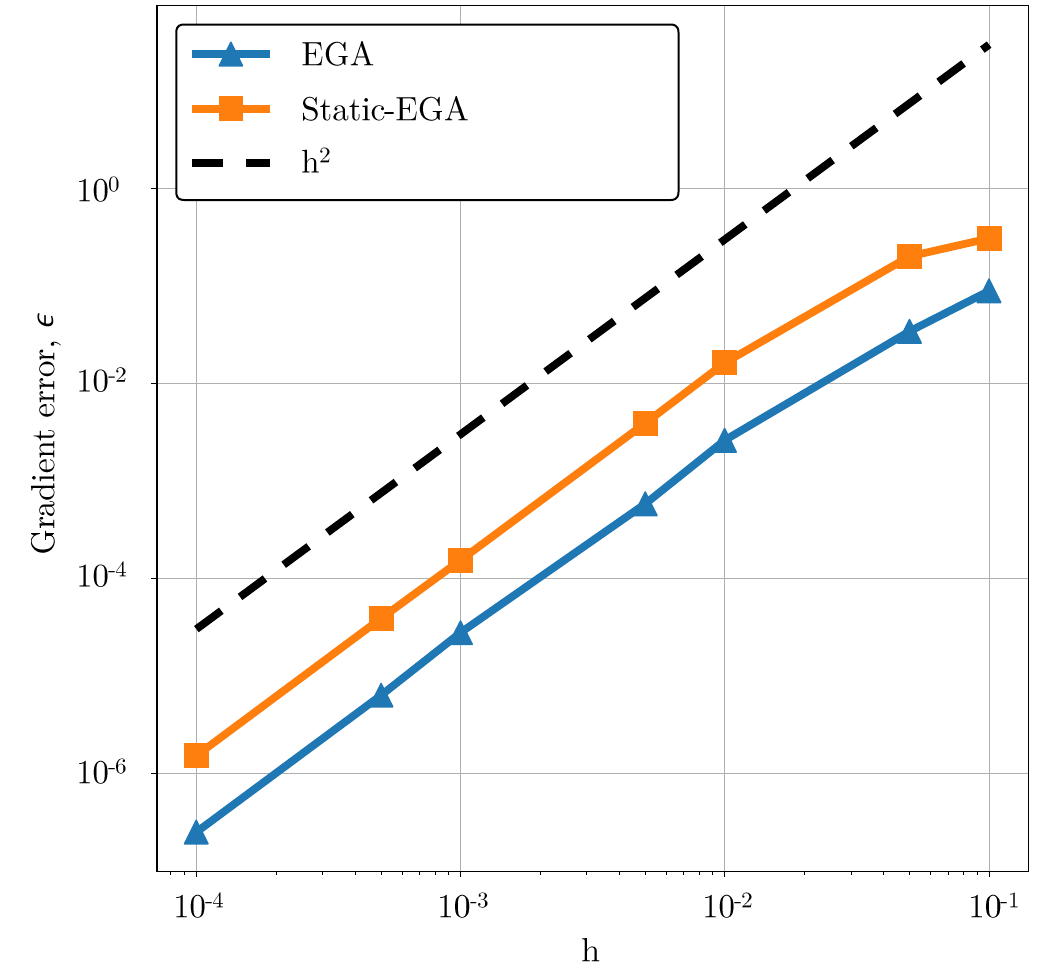}\label{fig:order_L63}
\caption{{{{ \bf  \em Order of convergence of the proposed Euler gradient approximations}. We provide experimental validation of the gradient based on both the EGA \eqref{eq:euler_grad_psitheorem} and Static-EGA \eqref{eq:gradpsi_identity_corollary1} formulas. The error is computed by comparing these gradients to the exact ones, returned using automatic differentiation of the solver $\Psi$. We recall that in the EGA, the Jacobian of the flow is computed exactly using automatic differentiation.}
}}
\label{fig:order_conv_L63}
\end{figure*}

Figure \ref{fig:order_conv_L63} shows the value of the gradient error for different values of $h$. Overall, the error of the proposed approximations behave as second order. This result validates numerically the development of theorem \ref{theorem:main_result_grad} and, importantly, the result of the Static-EGA (corollary \ref{corrol:gradpsi_identity_corollary} equation \eqref{eq:gradpsi_identity_corollary1}), that will be used in the following experiments.
 

\subsubsection{Sub-Model performance}
\label{exp:L63_pred}
In this experiment, we evaluate the performance of the proposed online learning techniques in deriving a relevant sub-model $\vect{M}_{\vect{\theta}}$. We assume that we are provided with a dataset of the true system in \eqref{eq:lorenz-63_vf}, $\mathcal{D}_{h} = \{(u^{\dagger}_{t_k +j{h_i}}, u^{\dagger}_{t_k})| \text{ with } k = 1 \dots N \text{ and } j = 1 \dots n\}$ and we aim at optimizing the parameters of the sub-model $\vect{M}_{\vect{\theta}}$ to minimize the cost of the online objective function \eqref{eq:online_loss}. We evaluate two of the proposed online learning schemes, the Static-EGA given in Eq. \eqref{eq:gradpsi_identity_corollary1} and the ETLM-EGA in which the Jacobian is approximated using an ensemble \eqref{eq:EnJacobian}. These schemes are compared to the simulations based on a sub-model that is calibrated using the exact gradient of the online objective. We also compare these models to the physical core, that we run without any correction.

\begin{figure*}
\centering
\includegraphics[width=1\textwidth]{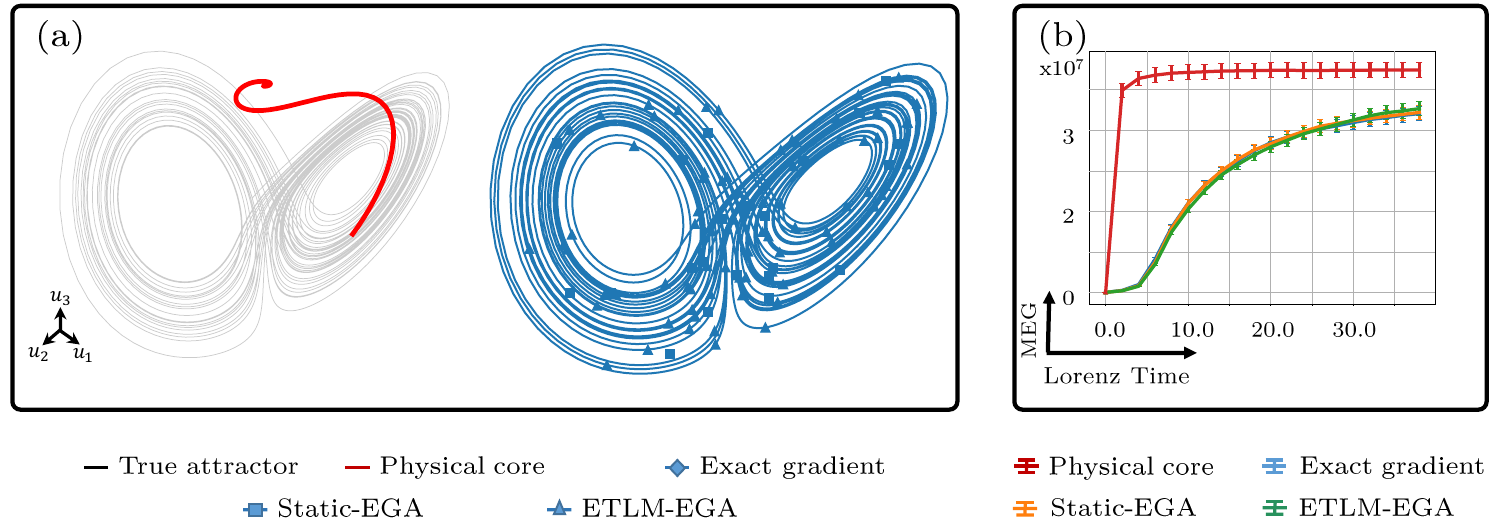}
\caption{{{{ \bf  \em Qualitative analysis of the tested models in the Lorenz 63 experiment}. (a) The attractor of the true Lorenz 63 model is compared to both the physical core (without the neural network correction term $\vect{M}_{\vect{\theta}}$) and to the hybrid models (with the neural network corrections). The sub-models are optimized online using stochastic gradient descent, and the gradients of the online function are computed either exactly (using automatic differentiation) or using some of the proposed approximations. (b) Mean error growth (MEG) of the tested models. The mean and standard deviation of the MEG are computed based on an ensemble of 20 trajectories issued from 5 different runs. The error bars are scaled by 1/20.}
}}
\label{fig:L63_phase_space}
\end{figure*}

A qualitative analysis of both the simulation and short-term forecasting performance of the tested models is given in Figure \ref{fig:L63_phase_space}. Overall, all the tested models are able to significantly improve the physical core, leading to a significant decrease in the forecasting error (as shown in panel (b)), while also reproducing the Lorenz 63 attractor (as highlighted in panel (a)). This finding is also validated through the computation of the Lyapunov spectrum and the Lyapunov dimension of the tested models, Table \ref{tab:fore_Lor_topo}. All the training schemes examined in this study yield simulations that closely align with the true underlying dynamics, which confirms the effectiveness of the proposed Euler approximations, for solving the online learning problem.

\begin{table}
\centering
\begin{tabular}{l*{2}c}
\toprule
Model &  Exponents & Dimension \\
\midrule \midrule 
True gradient & (0.90 -0.01 -14.57) $\pm$ (0.02 0.01 0.02) & 2.061 $\pm$ 0.001 \\
 midrule
Static approximation & (0.91 -0.01 -14.57) $\pm$ (0.02   0.01   0.02) & 2.061 $\pm$ 0.001 \\
\midrule
Ensemble approximation & (0.91 -0.01 -14.56) $\pm$ (0.02 0.01 0.02) & 2.061 $\pm$ 0.001 \\
\midrule
 Physical core & (-0.03 -4.99   -5.98) $\pm$ (0.01 1.43  1.42)& 0 \\
\bottomrule
\end{tabular}
\caption {{\bf  \em Simulation performance the data-driven models}: full Lyapunov spectrum and Lyapunov dimension of the tested models. The Lyapunov spectrum of the true Lorenz 63 system is (0.91, 0.0, -14.57) and it's dimension is estimated to be 2.064 \cite{Sprott_chaos}.}
\label{tab:fore_Lor_topo}
\end{table}

\subsection{Quasi-Geostrophic turbulence}
\subsubsection{Quasi-Geostrophic  dynamics}
In this second experiment, we consider Quasi-Geostrophic (QG) dynamics. QG theory is a workhorse to study geophysical fluid dynamics, relevant when the fluid follows an hydrostatic assumption and for which the Coriolis acceleration balances the horizontal pressure gradients. 

Over a doubly periodic $(x, y)$ square domain with length $L=2 \pi$, the dimensionless governing equations in the vorticity $(\omega_t)$ and streamfunction $(\psi_t)$ are:
\begin{subequations}
\label{eq:QG_equations_true}
\begin{align}
\frac{\partial \omega_t}{\partial t}+\mathcal{A}(\omega_t, \psi_t) & =\frac{1}{\operatorname{Re}} \nabla^2 \omega_t-f-r \omega_t \label{eq:equation_vorticity} \\
\nabla^2 \psi_t & =-\omega_t \label{eq:equation_stream_vorticity}
\end{align}
\end{subequations}
where, $\mathcal{A}(\omega_t, \psi_t)$ represents the nonlinear advection term:
$$
\mathcal{A}(\omega_t, \psi_t)=\frac{\partial \psi_t}{\partial y} \frac{\partial \omega_t}{\partial x}-\frac{\partial \psi_t}{\partial x} \frac{\partial \omega_t}{\partial y},
$$
and $f$ represents a deterministic forcing \cite{chandler2013invariant}:
$$
f(x, y)=k_f\left[\cos \left(k_f x\right)+\cos \left(k_f y\right)\right] \text {. }
$$
\subsubsection{Large eddy simulation setting}
In this experiment, we study the development of a sub-model that accounts for unresolved subgrid scale effects on the coarsened resolution of the QG equations \eqref{eq:QG_equations_true}. We are specifically interested in a Large Eddy Simulation (LES) setting in which the vorticity and streamfunction are filtered using a Gaussian filter \cite{sagaut2005large}, denoted by $\overline{(\cdot)}$. Applying this filter to Eqs. \eqref{eq:equation_vorticity}-\eqref{eq:equation_stream_vorticity} yields:
\begin{subequations}
\label{eq:QG_LES}
\begin{align}
\frac{\partial \bar{\omega}_t}{\partial t}+\mathcal{A}(\bar{\omega}_t, \bar{\psi}_t) & =\frac{1}{\operatorname{Re}} \nabla^2 \bar{\omega}_t-\bar{f}-r \bar{\omega}_t+\underbrace{\mathcal{A}(\bar{\omega}_t, \bar{\psi}_t)-\overline{\mathcal{A}(\omega_t, \psi_t)}}_{\Pi_t \approx \vect{M}_{\vect{\theta}}} \label{eq:equation_vorticity_LES} \\
\nabla^2 \bar{\psi}_t & =-\bar{\omega}_t \label{eq:equation_stream_vorticity_LES}
\end{align}
\end{subequations}
When compared to the direct numerical simulation (DNS) of equations \eqref{eq:equation_vorticity}-\eqref{eq:equation_stream_vorticity}, the LES can be solved at a coarser resolution. Yet, the term $\Pi_t$ encoding the Sub Grid Scale (SGS) variability requires a closure procedure. In this experiment, the proposed online learning techniques must recover a sub-model that accounts for this SGS term. This sub-model is a deep learning Convolutional Neural Network (CNN) $\vect{M}_{\vect{\theta}}(\bar{\psi}_t, \bar{\omega}_t)$ that takes as inputs both the vorticity and streamfunction fields $(\bar{\psi}_t, \bar{\omega}_t)$.

We use different random vorticity fields as initial conditions to generate 14 Direct Numerical Simulation (DNS) trajectories. These DNS data are then filtered to the resolution of the LES simulation and used as training, validation, and testing datasets. In this experiment, we compare the proposed online learning strategy with a static approximation of the Jacobian to both online learning with an exact gradient and offline learning schemes. These learning-based approaches are also compared to a standard, physics-based, dynamic Smagorinsky (DSMAG) parameterization \cite{germano1991dynamic}, in which the diffusion coefficient is constrained to be positive \cite{frezat2022posteriori} in order to avoid numerical instabilities related to energy backscattering \cite{guan2022stable,frezat2022posteriori}. 

\subsubsection{Offline analysis}
We first examine the accuracy of the deep learning sub-models in predicting the subgrid-scale term $\Pi_t$ for never-seen-before samples of $(\bar{\psi}_t, \bar{\omega}_t)$ within the testing set. We use a commonly used metric \cite{frezat2022posteriori,guan2023learning}, the correlation coefficient $c$ between the modeled $\vect{M}_{\vect{\theta}}(\bar{\psi}_t, \bar{\omega}_t)$ and true $\Pi_t$ SGS terms.

Table \ref{tab:corr_coef_offline} shows the correlation coefficients, averaged over two model runs and 1000 testing samples, for both the deep learning sub-models and the DSMAG baseline. Consistent with the previous findings \cite{frezat2022posteriori}, offline tests show that the data-driven SGS models substantially outperform DSMAG with a correlation coefficient $c$ above 0.8. We also validate, similarly to previous works \cite{frezat2022posteriori} that offline learning performs better on offline metrics than online learning-based models.

\begin{table}
    \centering
    \begin{tabular}{cccc}
    \hline
    
       Online, true gradient  & Online, Static-EGA & Offline & DSMAG\\
       \hline
       \hline
       $0.883 \pm 8.550\times 10^{-4} $   & $0.881 \pm  5.650 \times 10^{-3}$ & $0.930 \pm  2.009 \times 10^{-3}$ & $0.243$\\
       \hline
    \end{tabular}
\caption{{{{ \bf  \em Correlation coefficients between the predicted and true subgrid scale term}. }}}
    \label{tab:corr_coef_offline}
\end{table}

\subsubsection{Online analysis}
\begin{figure*}
\centering
\includegraphics[width=1\textwidth]{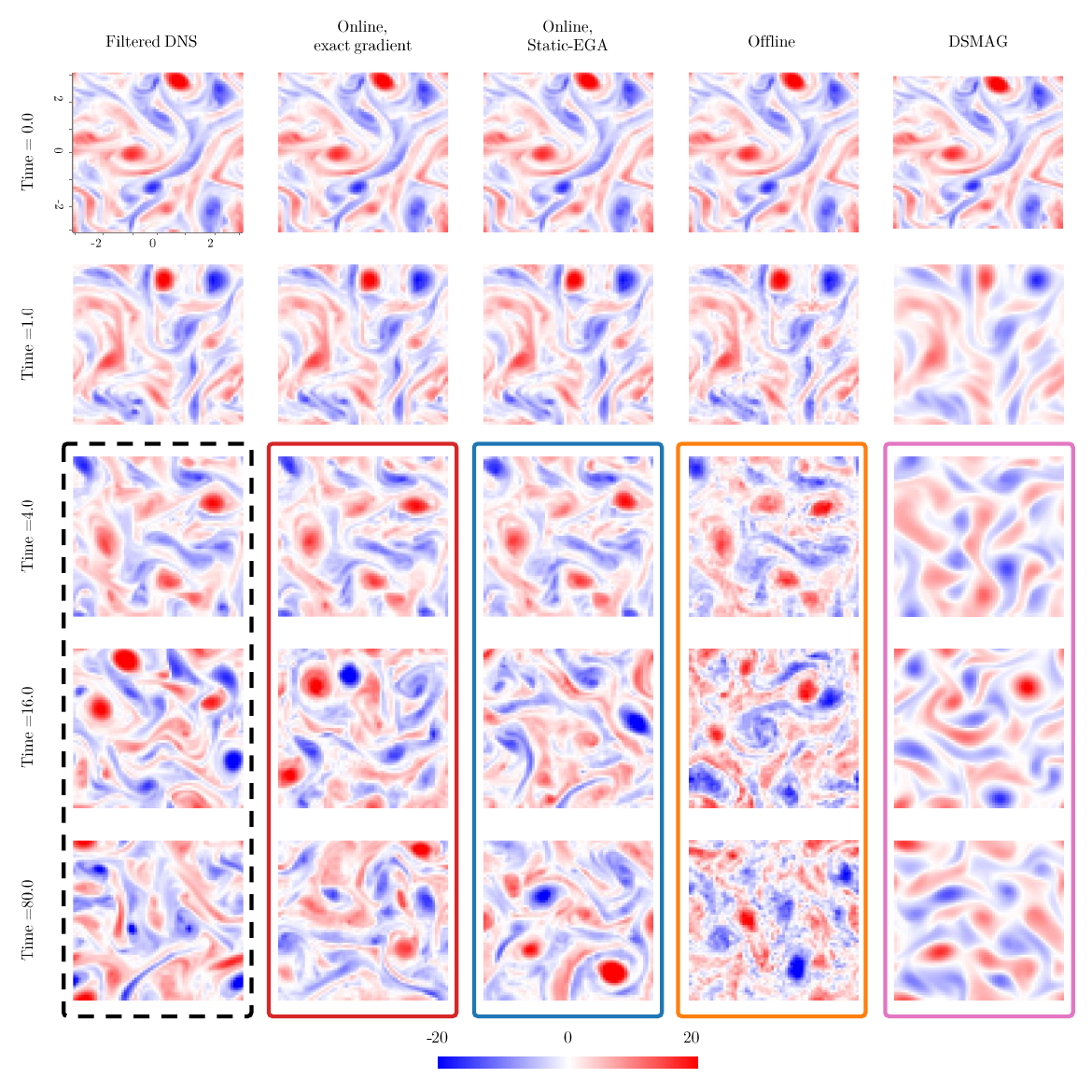}
\caption{{{{\bf \em Vorticity field simulation example for the different models}. Spectral analysis of the simulated trajectories, as well as the time series of the energy and enstrophy fields are given in Figure \ref{fig:pdf_spect_time_series}.}}}
\label{fig:Vorticity_plot}
\end{figure*}

Here, we evaluate the ability of the trained hybrid models to reproduce the dynamics of the filtered DNS simulation. Figure \ref{fig:Vorticity_plot} shows a simulation example from an initial condition in the test set. A visual analysis reveals that the DSMAG scheme (purple panel in Fig. \ref{fig:Vorticity_plot}) smooths out fine-scale structures. This model is intrinsically built on a diffusion assumption which enables the system to sustain large-scale variability at the cost of excessively smoothing small-scale features. Although deep learning-based sub-models calibrated offline demonstrate superior offline performance, indicated by a high correlation coefficient (refer to Table \ref{tab:corr_coef_offline}), their coupling with the solver results in an unphysical behavior within the coupled hybrid dynamical system. Specifically, the online performance of the offline sub-model shows, in the orange panel in Fig. \ref{fig:Vorticity_plot}, sustained high-resolution variability, that is not present in the filtered DNS. 
The results of online learning with the exact gradient show more realistic flows, that are visually comparable to the the filtered DNS. The blue panel in Figure \ref{fig:Vorticity_plot}, demonstrates that the proposed Euler approximation of the online learning problem also provides a flow close to the filtered DNS, without relying on a differentiable solver.

\begin{figure*}
\centering
\includegraphics[width=1\textwidth]{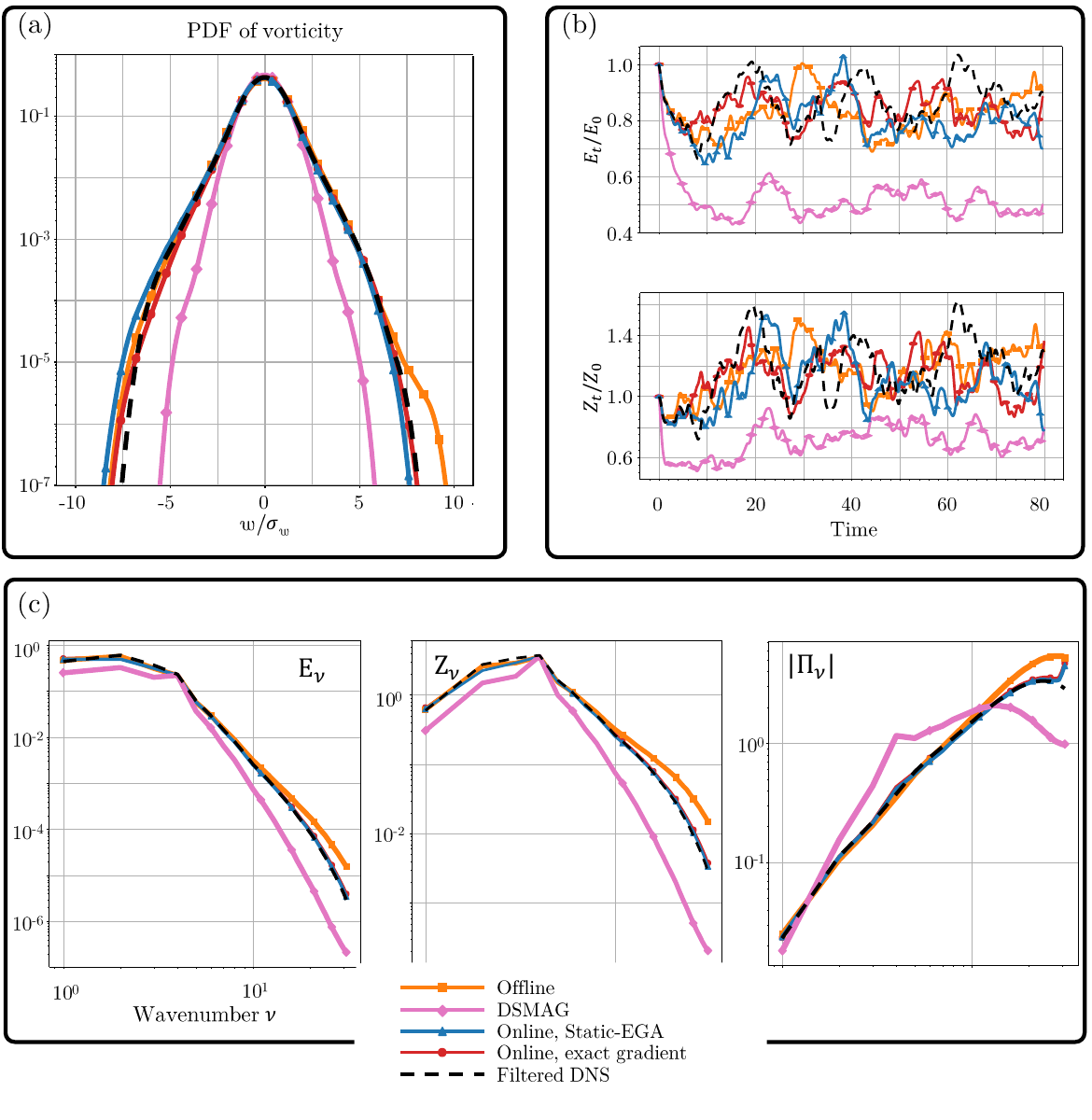}
\caption{{{{ \bf  \em Statistical evaluation of the different models}. (a) Probability density function of the vorticity field (b) Time evolution of the kinetic energy $E_t = \frac{1}{2}<\overline{\psi}_t \overline{\omega}_t>$ and enstrophy $Z_t = \frac{1}{2}<\overline{\omega}_t^2>$ normalized by the energy and enstrophy of the initial condition (of the filtered DNS) $E_0$ and $Z_0$ respectively. (c) Time averaged kinetic energy spectra $E_{\nu}$, enstrophy spectra $E_{\nu}$ and power spectrum of the SGS term $|\Pi_{\nu}|$. The PDF is computed using a kernel density estimator.}
}}
\label{fig:pdf_spect_time_series}
\end{figure*}

We draw similar conclusions from the statistical properties of the simulated flows reported in Figure \ref{fig:pdf_spect_time_series}. Overall, the analysis of the Probability Density Function (PDF) of the vorticity field in Fig. \ref{fig:pdf_spect_time_series} confirms that online learning schemes display the closest statistical properties to the filtered DNS field. The DSMAG sub-model does not reproduce the extreme events of both positive and negative vorticity, and the offline learning-based model exhibits a skewed distribution tail, which is predominantly biased towards positive vorticity values. Likely, the time series of kinetic energy $E_t$ and enstrophy $Z_t$ demonstrate that the DSMAG model, characterized by excessive diffusivity and the absence of backscattering, results in a substantial reduction in both $Z_t$ and $E_t$. The analysis of the time-averaged spectra also clearly highlights the robustness of the sub-models that are trained online. Both the one optimized with the exact gradient and the proposed approximation accurately reproduce small and large scales of the flow.

\subsubsection{Fine-tuning offline sub-models}
\begin{figure*}
\centering
\includegraphics[width=1\textwidth]{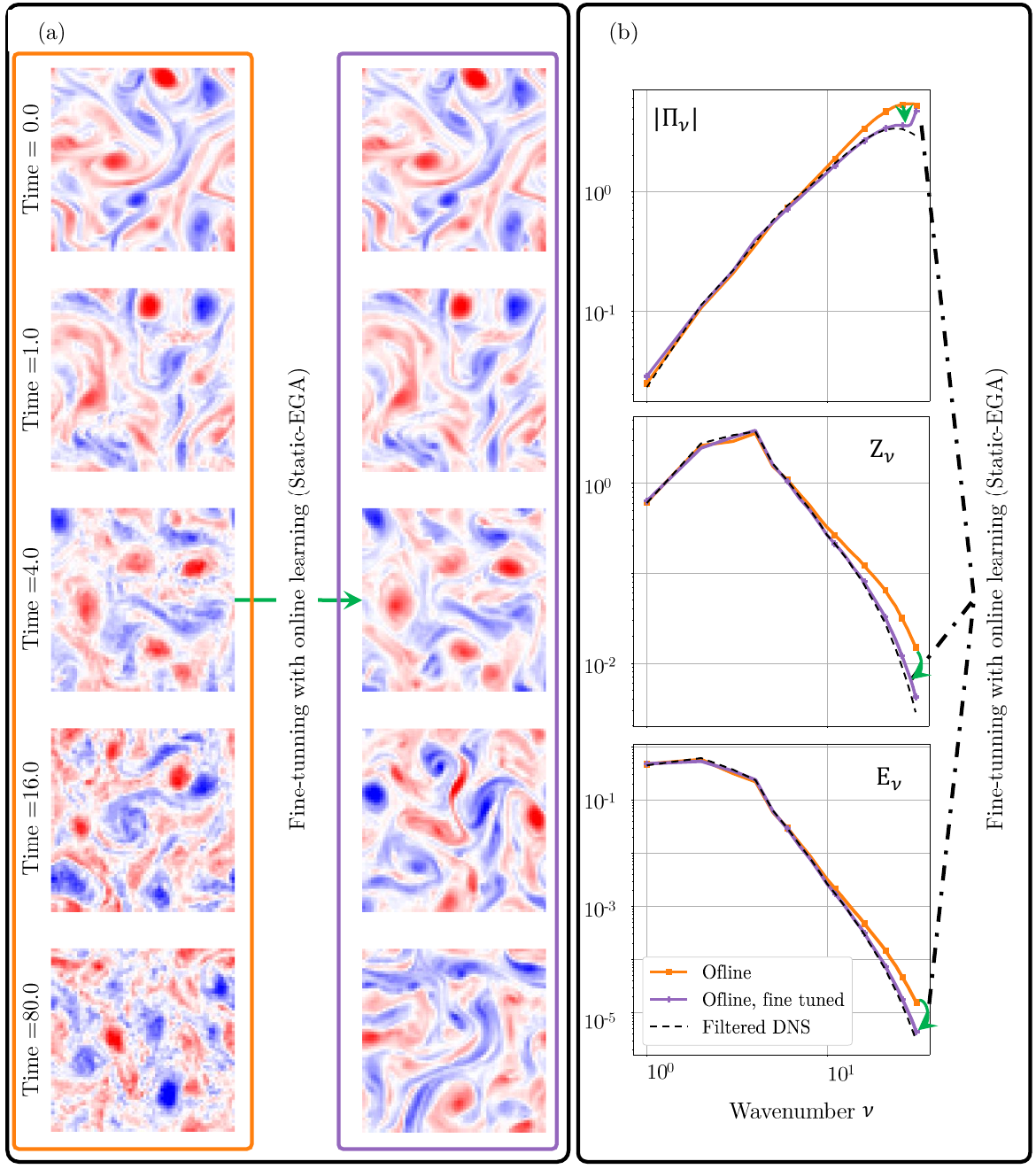}
\caption{{{{ \bf  \em Vorticity field and time-averaged spectrums of the offline model before and after fine-tunning}. (a) The same offline model that leads to an unphysical vorticity field in Figure \ref{fig:Vorticity_plot} is fine-tuned using the proposed static approximation of the online learning problem, leading to a realistic vorticity simulation. (b) Time-averaged kinetic energy spectra $E_{\nu}$, enstrophy spectra $Z_{\nu}$ and power spectrum of the SGS term $|\Pi_{\nu}|$ of the offline model before and after the online fine-tuning.}
}}
\label{fig:Vorticity_plot_fine_tuning}
\end{figure*}

The proposed online optimization approach can be beneficial in situations where a sub-model, calibrated offline, displays a nonphysical or unstable behavior when coupled with the solver. Such a sub-model can be significantly improved by fine-tuning its parameters using an online optimization scheme and our proposed gradient approximation allows achieving this fine-tuning step, without requiring access to the exact gradient of the solver. Figure \ref{fig:Vorticity_plot_fine_tuning} displays the improvements of the offline model discussed above when fine-tuned (for two epochs) using the proposed static approximation of the online learning. A visual inspection of the vorticity field in Figure \ref{fig:Vorticity_plot_fine_tuning}, shows how this fine-tuning step successfully removes the unphysical behavior of the offline model. This fine-tuning step also brings significant improvement to the time-averaged spectra, shown in \ref{fig:Vorticity_plot_fine_tuning}, now aligning closely with the filtered DNS spectra.

\section{Outlook}
\label{sec:scoop_lims}

With these promising results, we foresee future investigations to scale such online learning schemes to higher-complexity simulation frameworks. Challenges include accounting in the proposed approach for multiple sub-models that might not be additive. 
Extensions to stochastic sub-models are also appealing to address probabilistic forecasting and data assimilation challenges. Finally, we discuss the decomposition of the learned sub-models into a combination of interpretable physical processes.

\subsection{Extension to multiple sub-models}
The sub-model $\vect{M}_{\vect{\theta}}$ in \eqref{eq:PDE_disc} may often account for various unresolved processes (e.g., passive and active tracer transport, biogeochemistry, convection) which are typically governed by different underlying dynamics. To calibrate a number of $q$ sub-models, and assuming that these models are additive, equation \eqref{eq:PDE_disc} becomes:
\begin{equation}
  \left\{
    \begin{array}{l l}
      \Dot{\vect{u}}_t &= \vect{F}(\vect{u}_t) + \sum_{i = 1}^{q} \vect{M}^i_{\vect{\theta}}(\vect{u}_t)\\
      \vect{u}(0) &= \vect{u}_0
    \end{array}
  \right.
  \label{eq:PDE_disc_several_sub-models}
\end{equation}
Jointly calibrating these sub-models using an online learning methodology should certainly be considered with care to avoid mixing the individual contribution of every sub-model. To circumvent such a difficulty, it is necessary to define and include offline costs for each sub-model as a regularization of the online learning objective function. For every sub-model $\vect{M}^i_{\vect{\theta}}$, we need to define, as discussed in the offline learning section \ref{seq:offline_learning}, a reference dataset $\vect{R}^i_t$ that represents an ideal response to the input $\vect{u}_t$. The new online cost function that takes into account these offline regularizations can be written as:
\begin{equation*}
\mathcal{J}_{combined} = \mathcal{J}(\vect{g}^{\dagger}({u}^{\dagger}_{t+nh}), \vect{g}(\Psi^n(\vect{u}_{t})),\vect{\theta}) +\sum_{i = 1}^{q} \mathcal{Q}^i(\vect{R}^i_{t}, \vect{M}^i_{\vect{\theta}},\vect{\theta}) 
\end{equation*}
where $\mathcal{J}(\cdot)$ is the online cost function. Minimizing each individual regularization term within this learning methodology ensures that every sub-model accurately represents a specified underlying process. Simultaneously, the online objective function guarantees the correct interaction among these sub-models and with the physical core $\vect{F}$.

\subsection{Stochastic hybrid models}
In the present development, interactions between the resolved flow component and the unresolved components are formulated as correction terms that are constructed based on the resolved components of the flow. Likely, we are missing several degrees of freedom that represent the independent or generic fine-scale variability. This missing variability generates uncertainty. Taking into account and modeling this uncertainty is mandatory in applications that require probabilistic forecasting, such as data assimilation \cite{zhen2023physically}. Models that encode uncertainty can consistently define Stochastic versions of the original equations and can ensure that certain quantities, such as integrals of functions over a spatial volume like momentum, mass, matter, and energy, are conserved at every time step \cite{memin2014fluid,holm2015variational}. The proposed online learning methodology can allow for the calibration of such stochastic sub-models for various configurations regarding the model class and the objective function formulation. 

\subsection{Beyond additive sub-models}
The proposed Euler gradient approximations are based on the assumption that the sub-model in \eqref{eq:PDE_disc} is additive. Precisely, this assumption allows us to express the gradient of the Euler solver, for a given initial condition, in terms of the gradient of the sub-model. The latter is then easily evaluated using automatic differentiation. Additive sub-models are commonly found in physical simulations, enabling the proposed framework to cover a wide range of case studies. Yet, the proposed methodology can be extended to non-additive sub-models by converting them into additive ones. This conversion can be achieved by explicitly coding the non-additive physical contribution into the sub-model $\vect{M}_{\theta}$.

\subsection{Towards explainable sub-models}
Understanding the mechanisms responsible for the success of the deep learning model will improve the reliability of such representations. Interpreting deep learning sub-models might also guide the analysis of the interaction between the large-scale physical core and the heuristic processes that are represented by the sub-model. In particular, the sub-model $\vect{M}_{\theta}$ can be decomposed into a combination of known expected physical contributions. In the context of subgrid-scale modeling, these contributions might include the dissipative effects acting on the smallest scales, and some advection corrections and backscattering effects acting on the transport of the resolved quantities. New decompositions can then be considered to, for instance, more effectively include the dissipation explicitly and/or anticipated advection corrections in $\vect{F}$. 

In addition to subgrid-scale modeling and interpretation, extracting other sources of information including errors, due for instance to the presence of systematic errors, biases, or overall incompatibility of $\vect{F}$ in the modeling of some observable $g(\cdot)$, could also be addressed. It is a challenging problem and must require capabilities to disentangle all the contributions from the sub-model $\vect{M}_{\vect{\theta}}$, while taking into account the possible errors and the overall (positive or negative) impact of $\vect{F}$ into the modeling of $g(\cdot)$. For future investigations, we believe that valuable insights on this problem can be obtained by developing ensembles of simulations, performed at various spatial scales, to subsequently analyze the inferred different sub-models.

\section{conclusion}
\label{seq:conc}
Nowadays, the high-fidelity simulation and forecasting of physical phenomena rely on dynamical cores derived from well-understood physical principles, along with sub-models that approximate the impacts of certain phenomena that are either unknown or too expensive to be resolved explicitly. Recently, deep learning techniques brought attention and potential to the definition and calibration of these sub-models. In particular, online learning strategies provide appealing solutions to improve numerical models and make them closer to actual observations.  

In this work, we develop the EGA, an easy-to-use workflow that allows for the online training of sub-models of hybrid numerical modeling systems. It bypasses the differentiability bottleneck of the physical models and converges to the exact gradients as the time step tends to zero. We stress the robustness and efficiency of the proposed online learning scheme on realistic case studies, including Quasi-Geostrophic dynamics. Overall, we report significant improvements when compared to standard offline learning schemes and achieve a performance that is similar to solving the exact online learning problem.

Our future work will explore the proposed methodology for large-scale realistic models such as the ones used in atmosphere and ocean simulations. Besides algorithm developments, it will require technical efforts to synchronize the PDE solvers, usually written in high-performance languages such as FORTRAN, to the deep learning-based sub-models that are usually in languages and packages that support automatic differentiation (Pytorch, Jax, Tensorflow). We will particularly focus on the end-to-end calibration of hybrid systems with observation-driven constraints and expect to improve forecasting performance when compared to standard physical models and to full data-driven forecasting surrogates as the ones developed in \cite{pathak2022fourcastnet,lam2022graphcast,dueben2022challenges,chen2023fengwu, bi2023accurate,nguyen2023climax}.

\section*{Acknowledgement}
This work was supported by the ERC Synergy project 856408-STUOD, Labex Cominlabs (grant SEACS), CNES (grant OSTST-MANATEE), Microsoft (AI EU Ocean awards), ANR Projects Melody and OceaniX. It benefited from HPC and GPU resources from Azure (Microsoft EU Ocean awards) and GENCI-IDRIS (Grant 2020-101030).

\bibliographystyle{unsrtnat}
\bibliography{references} 

\newpage
\appendix

\section{Proof of the proposition \ref{prop:psi_as_euler}}
Since the solver $\Psi$ is of order $p$, and the explicit Euler scheme has order $1$ it is trivial to prove proposition \eqref{prop:psi_as_euler} from the Taylor expansion of $\Psi$. We include the proof here for completeness. We can write as as $h$ approaches zero:
\begin{equation}
\label{eq:proof_prop_}
\begin{aligned}
\vect{u}_{t + h} &= \Psi(\vect{u}_t) \\
 &= \vect{u}_t + \sum_{k=1}^{p} h^k\frac{1}{k!}(\vect{F}(\vect{u}_t) + \vect{M}_{\vect{\theta}}(\vect{u}_t))^{k-1} + O(h^{p+1}) \\
 &= \underbrace{\vect{u}_t + h(\vect{F}(\vect{u}_t) + \vect{M}_{\vect{\theta}}(\vect{u}_t))}_{\Psi_E(\vect{u}_t)} +  \sum_{k=2}^{p} h^k\frac{1}{k!}(\vect{F}(\vect{u}_t) + \vect{M}_{\vect{\theta}}(\vect{u}_t))^{k-1} + O(h^{p+1}) \\ 
 &= \Psi_E(\vect{u}_t) + \underbrace{\sum_{k=2}^{p} h^k\frac{1}{k!}(\vect{F}(\vect{u}_t) + \vect{M}_{\vect{\theta}}(\vect{u}_t))^{k-1}}_{<C h^2} + O(h^{p+1})\\
  &= \Psi_E(\vect{u}_t) + O(h^{2})
\end{aligned}
\end{equation}
The second part of the proposition can be proven similarly by replacing $\vect{u}_t$ by $\Psi^{n-1}(\vect{u}_t)$.

\section{Proof of theorem \ref{theorem:main_result_grad}}
Notice that for every $n$, the following holds:
\begin{equation}
\label{eq:gradient_of_psi_elementaire_noO}
\begin{aligned}
\frac{\partial}{\partial\vect{\theta}}  {\Psi^n}(\vect{u}_t) &= \frac{\partial}{\partial\vect{\theta}}  {\Psi} \circ {\Psi^{n-1}}(\vect{u}_t) \\
&= \underbrace{\frac{\partial \Psi({\Psi^{n-1}}(\vect{u}_t))}{\partial {\Psi^{n-1}}(\vect{u}_t)}   \frac{\partial {\Psi^{n-1}}(\vect{u}_t)}{\partial\vect{\theta}}}_{\text{Variation due to the initial condition}} + \underbrace{\frac{\partial \Psi({\Psi^{n-1}}(\vect{u}_t))}{\partial\vect{\theta}}}_{\text{Variation of the gradient of the solver given the initial condition}} 
\end{aligned}
\end{equation}

We use \eqref{eq:gradient_of_psi_elementaire_noO} to construct the following:
\begin{equation}
\label{eq:euler_grad_psitheorem_psi}
\begin{aligned}
\frac{\partial}{\partial\vect{\theta}}  {\Psi^n}(\vect{u}_{t}) = \sum_{j = 1}^{j = n-1} (\prod_{i = 1}^{i = n-j} \frac{\partial \Psi(\Psi^{n-i}(\vect{u}_{t}))}{\partial \Psi^{n-i}(\vect{u}_{t})}) \frac{\partial}{\partial\vect{\theta}} \Psi(\Psi^{j -1}(\vect{u}_{t})) + \frac{\partial}{\partial\vect{\theta}} \Psi(\Psi^{n-1}(\vect{u}_{t}))
\end{aligned}
\end{equation}
where all the derivatives with respect to $\theta$ are taken assuming that the initial condition is fixed. The proof of \eqref{eq:euler_grad_psitheorem_psi} is conducted by recurrence, and is given in appendix \ref{seq:prof_recu}.

To prove theorem \ref{theorem:main_result_grad}, we simply replace in \eqref{eq:euler_grad_psitheorem_psi} the solver $\Psi$ by its Euler approximation given in the proposition \ref{prop:psi_as_euler}:
\begin{equation}
\label{eq:gradpsiasgradpsue}
\begin{aligned}
\frac{\partial}{\partial\vect{\theta}}  {\Psi^n}(\vect{u}_{t}) = \sum_{j = 1}^{j = n-1} (\prod_{i = 1}^{i = n-j} \frac{\partial \Psi(\Psi^{n-i}(\vect{u}_{t}))}{\partial \Psi^{n-i}(\vect{u}_{t})}) \frac{\partial}{\partial\vect{\theta}} (\Psi_E(\Psi^{j -1}(\vect{u}_{t})) + O(h^2)) + \frac{\partial}{\partial\vect{\theta}}(\Psi_E(\Psi^{n-1}(\vect{u}_{t})) + O(h^2)) 
\end{aligned}
\end{equation}
if we develop the expression of the explicit Euler solver, we have:
\begin{equation}
\label{eq:euler_grad_xx}
\begin{aligned}
\frac{\partial}{\partial\vect{\theta}}  {\Psi^n}(\vect{u}_{t}) &= \sum_{j = 1}^{j = n-1} (\prod_{i = 1}^{i = n-j} \frac{\partial \Psi(\Psi^{n-i}(\vect{u}_{t}))}{\partial \Psi^{n-i}(\vect{u}_{t})}) \frac{\partial}{\partial\vect{\theta}} (\Psi^{j -1}(\vect{u}_{t}) + h (\vect{F}(\Psi^{j -1}(\vect{u}_{t})) + \vect{M}_{\vect{\theta}}(\Psi^{j -1}(\vect{u}_{t}))) +O(h^2)) \\
&+ \frac{\partial}{\partial\vect{\theta}} (\Psi^{n -1}(\vect{u}_{t}) + h (\vect{F}(\Psi^{n -1}(\vect{u}_{t})) + \vect{M}_{\vect{\theta}}(\Psi^{n -1}(\vect{u}_{t})))+O(h^2))\\
&= \sum_{j = 1}^{j = n-1} (\prod_{i = 1}^{i = n-j} \frac{\partial \Psi(\Psi^{n-i}(\vect{u}_{t}))}{\partial \Psi^{n-i}(\vect{u}_{t})}) h \frac{\partial}{\partial\vect{\theta}} \vect{M}_{\vect{\theta}}(\Psi^{j -1}(\vect{u}_{t})) +\sum_{j = 1}^{j = n-1}  (\prod_{i = 1}^{i = n-j} \frac{\partial \Psi(\Psi^{n-i}(\vect{u}_{t}))}{\partial \Psi^{n-i}(\vect{u}_{t})})O(h^2) \\
&+ h \frac{\partial}{\partial\vect{\theta}} \vect{M}_{\vect{\theta}}(\Psi^{n -1}(\vect{u}_{t}))+O(h^2)\\
&= \sum_{j = 1}^{j = n-1} (\prod_{i = 1}^{i = n-j} \frac{\partial \Psi(\Psi^{n-i}(\vect{u}_{t}))}{\partial \Psi^{n-i}(\vect{u}_{t})}) h \frac{\partial}{\partial\vect{\theta}} \vect{M}_{\vect{\theta}}(\Psi^{j -1}(\vect{u}_{t}))+ h \frac{\partial}{\partial\vect{\theta}} \vect{M}_{\vect{\theta}}(\Psi^{n -1}(\vect{u}_{t}))+nO(h^2)
\end{aligned}
\end{equation}
The leading order term in $\sum_{j = 1}^{j = n-1}  (\prod_{i = 1}^{i = n-j} \frac{\partial \Psi(\Psi^{n-i}(\vect{u}_{t}))}{\partial \Psi^{n-i}(\vect{u}_{t})})O(h^2)$ is $O(h^2)$. Furthermore, and since $n$ is fixed, $nO(h^2)$ is simply equal to $O(h^2)$. Reporting this in equation \eqref{eq:euler_grad_xx} completes the proof as follows:
\begin{equation}
\begin{aligned}
\frac{\partial}{\partial\vect{\theta}}  {\Psi^n}(\vect{u}_{t}) &= \sum_{j = 1}^{j = n-1} (\prod_{i = 1}^{i = n-j} \frac{\partial \Psi(\Psi^{n-i}(\vect{u}_{t}))}{\partial \Psi^{n-i}(\vect{u}_{t})}) h \frac{\partial}{\partial\vect{\theta}} \vect{M}_{\vect{\theta}}(\Psi^{j -1}(\vect{u}_{t}))+ h \frac{\partial}{\partial\vect{\theta}} \vect{M}_{\vect{\theta}}(\Psi^{n -1}(\vect{u}_{t}))+O(h^2)
\end{aligned}
\end{equation}

\section{Proof of the corollary \ref{corrol:approx_grad_h}}
We prove corollary \eqref{corrol:approx_grad_h}, we start from \eqref{eq:euler_grad_xx} and replace $n$ by $\frac{t_f-t_0}{h}$. This makes the convergence linear in $h$:
\begin{equation}
\begin{aligned}
\frac{\partial}{\partial\vect{\theta}}  {\Psi^n}(\vect{u}_{t}) &= \sum_{j = 1}^{j = n-1} (\prod_{i = 1}^{i = n-j} \frac{\partial \Psi(\Psi^{n-i}(\vect{u}_{t}))}{\partial \Psi^{n-i}(\vect{u}_{t})}) h\frac{\partial}{\partial\vect{\theta}} \vect{M}_{\vect{\theta}}(\Psi^{j -1}(\vect{u}_{t}))+ h \frac{\partial}{\partial\vect{\theta}} \vect{M}_{\vect{\theta}}(\Psi^{n -1}(\vect{u}_{t}))+O(h)
\end{aligned}
\end{equation}

\section{Proof of the Corollary \ref{corrol:gradpsi_identity_corollary}}
In order to prove corollary \ref{corrol:gradpsi_identity_corollary}, we write the Jacobian of the taylor expansion of the solver $\Psi$ and we keep the zero order term.
\begin{equation}
\label{eq:proof_prop_zero_order}
\begin{aligned}
\frac{\partial}{\partial \Psi(\vect{u}_t)} \Psi(\Psi(\vect{u}_t)) &= \frac{\partial}{\partial \Psi(\vect{u}_t)} (\Psi(\vect{u}_t) + \sum_{k=1}^{p} h^k\frac{1}{k!}(\vect{F}(\Psi(\vect{u}_t)) + \vect{M}_{\vect{\theta}}(\Psi(\vect{u}_t)))^{k-1} + O(h^{p+1})) \\
 &= \textbf{I} + \frac{\partial}{\partial \Psi(\vect{u}_t)} \sum_{k=1}^{p} h^k\frac{1}{k!}(\vect{F}(\Psi(\vect{u}_t)) + \vect{M}_{\vect{\theta}}(\Psi(\vect{u}_t)))^{k-1} + O(h^{p+1})\\
  &= \textbf{I} + O(h)\\
\end{aligned}
\end{equation}

If we replace the Jacobian in \eqref{eq:euler_grad_psitheorem} by the approximation in \eqref{eq:proof_prop_zero_order}:
\begin{equation}
\begin{aligned}
\frac{\partial}{\partial\vect{\theta}}  {\Psi^n}(\vect{u}_{t}) &= \sum_{j = 1}^{j = n-1} (\prod_{i = 1}^{i = n-j} \textbf{I} + O(h) ) h \frac{\partial}{\partial\vect{\theta}} \vect{M}_{\vect{\theta}}(\Psi^{j -1}(\vect{u}_{t}))+ h \frac{\partial}{\partial \vect{\theta}} \vect{M}_{\vect{\theta}}(\Psi^{n -1}(\vect{u}_{t}))+O(h^2)\\
&= \sum_{j = 1}^{j = n-1} (\textbf{I} + O(h) ) h \frac{\partial}{\partial \vect{\theta}} \vect{M}_{\vect{\theta}}(\Psi^{j -1}(\vect{u}_{t}))+ h \frac{\partial}{\partial \vect{\theta}} \vect{M}_{\vect{\theta}}(\Psi^{n -1}(\vect{u}_{t}))+O(h^2)\\
&= \sum_{j = 1}^{j = n-1} h \frac{\partial}{\partial \vect{\theta}} \vect{M}_{\vect{\theta}}(\Psi^{j -1}(\vect{u}_{t}))+ h \frac{\partial}{\partial \vect{\theta}} \vect{M}_{\vect{\theta}}(\Psi^{n -1}(\vect{u}_{t}))+\sum_{j = 1}^{j = n-1} O(h)h \frac{\partial}{\partial \vect{\theta}} \vect{M}_{\vect{\theta}}(\Psi^{j -1}(\vect{u}_{t})) + O(h^2)\\
&= \sum_{j = 1}^{j = n-1} h \frac{\partial}{\partial \vect{\theta}} \vect{M}_{\vect{\theta}}(\Psi^{j -1}(\vect{u}_{t}))+ h \frac{\partial}{\partial \vect{\theta}} \vect{M}_{\vect{\theta}}(\Psi^{n -1}(\vect{u}_{t}))+ nO(h^2)\\
&= \sum_{j = 1}^{j = n} h \frac{\partial}{\partial \vect{\theta}} \vect{M}_{\vect{\theta}}(\Psi^{j -1}(\vect{u}_{t}))+ O(h^2)
\end{aligned}
\end{equation}
The second part of the corollary, {\em i.e.} assuming that $n = \frac{t_f-t_0}{h}$ can be proven similarly. 

\section{Proof of the Corollary \ref{corrol:TLM}}
\begin{equation}
\label{eq:proof_coroll_TLM}
\begin{aligned}
\frac{\partial}{\partial \Psi(\vect{u}_t)} \Psi(\Psi(\vect{u}_t)) &= \frac{\partial}{\partial \Psi(\vect{u}_t)} (\Psi(\vect{u}_t) + \sum_{k=1}^{p} h^k\frac{1}{k!}(\vect{F}(\Psi(\vect{u}_t)) + \vect{M}_{\vect{\theta}}(\Psi(\vect{u}_t)))^{k-1} + O(h^{p+1})) \\
&= \frac{\partial}{\partial \Psi(\vect{u}_t)} (\Psi(\vect{u}_t) + \sum_{k=1}^{p} h^k\frac{1}{k!}(\vect{F}(\Psi(\vect{u}_t)))^{k-1} +\sum_{k=1}^{p} h^k\frac{1}{k!}(\vect{M}_{\vect{\theta}}(\Psi(\vect{u}_t)))^{k-1} + O(h^{p+1}))\\
&= \underbrace{\frac{\partial}{\partial \Psi(\vect{u}_t)} (\Psi(\vect{u}_t) + \sum_{k=1}^{p} h^k\frac{1}{k!}(\vect{F}(\Psi(\vect{u}_t)))^{k-1}+ O(h^{p+1}))}_{\frac{\partial}{\partial \Psi(\vect{u}_t)} \Psi_o(\Psi(\vect{u}_t))} +\frac{\partial}{\partial \Psi(\vect{u}_t)} \sum_{k=1}^{p} h^k\frac{1}{k!}(\vect{M}_{\vect{\theta}}(\Psi(\vect{u}_t)))^{k-1} \\
&=\frac{\partial}{\partial \Psi(\vect{u}_t)} \Psi_o(\Psi(\vect{u}_t)) + \frac{\partial}{\partial \Psi(\vect{u}_t)} \sum_{k=1}^{p} h^k\frac{1}{k!}(\vect{M}_{\vect{\theta}}(\Psi(\vect{u}_t)))^{k-1}\\
&=TLM(\Psi(\vect{u}_t)) +\frac{\partial}{\partial \Psi(\vect{u}_t)} \sum_{k=1}^{p} h^k\frac{1}{k!}(\vect{M}_{\vect{\theta}}(\Psi(\vect{u}_t)))^{k-1}
\end{aligned}
\end{equation}

\section{Algorithms of the proposed online learning methodology}
\label{sec:algos}
\subsection{Gradient evaluation using composable function transforms}
A direct evaluation of \eqref{eq:grad_online_autodiff} can be based on composable function transforms of modern languages such as JAX \cite{bradbury2018jax} or PYTORCH (based on the functorch tool). These tools allow to evaluate vector valued gradients (not only vector Jacobian products), and it can be adapted to the computation of \eqref{eq:grad_online_autodiff}. Algorithm \ref{algo1} highlights how this can be achieved.
\begin{algorithm}[h]
\caption{Gradient computation based on composable function transforms}
\label{algo1}
\begin{algorithmic}
\Require
    \Statex $\Psi$: Non-differentiable solver of the hybrid system \eqref{eq:flow_discretized_equation}
    \Statex $\vect{M}_{\vect{\theta}}$: Deep learning based sub-model
    \Statex $\vect{u}_t$: Initial condition
    \Statex $n, h$: Number of simulation steps and time step
    \Statex $\mathcal{J}_{online}$ : Online cost function
    \Statex $l$ : Approximation scheme for the Jacobian of the flow
    
\Ensure    
\LineComment{\textit{Iterate through the solver $\Psi$}}
\For{$j \gets 1$ \textbf{to} $n$}

    $\vect{u}_{t + jh} = \Psi^j(\vect{u}_{t})$
\EndFor

\LineComment{\textit{Precompute the Jacobians $\vect{J}_{j,l}$}}
\For{$j \gets 1$ \textbf{to} $n-1$}

    $\vect{J}_{j,l} = \prod_{i = 1}^{i = n-j} \nicefrac{\partial \Psi(\Psi^{n-i}(\vect{u}_{t}))}{\partial \Psi^{n-i}(\vect{u}_{t})}$
\EndFor

\LineComment{\textit{Compute the vector valued gradients of the sub-model}}
\For{$j \gets 1$ \textbf{to} $n$}

    $\nicefrac{\partial \vect{M}_{\vect{\theta}}(\Psi^{j-1}(\vect{u}_{t}))}{\partial\vect{\theta}}$ \Comment{\textit{This can be done, for example, using functorch}}
\EndFor

\LineComment{\textit{Compute the Euler approximation of the gradient of the solver}}

\State $\vect{A}_{l,p} = \sum_{j = 1}^{j = n-1} \vect{J}_{j,l} h \nicefrac{\partial}{\partial\vect{\theta}} \vect{M}_{\vect{\theta}}(\Psi^{j -1}(\vect{u}_{t}))+ h \nicefrac{\partial}{\partial\vect{\theta}} \vect{M}_{\vect{\theta}}(\Psi^{n -1}(\vect{u}_{t})))$

\LineComment{\textit{Compute the remaining gradients and evaluate the gradient of the online cost }}

\State  $\vect{v} = \nicefrac{\partial Q(\cdot,\cdot,\theta)}{\partial\vect{\theta}}$ 

\State  $\vect{w} = \nicefrac{\partial Q(\cdot,\vect{g}(\Psi^n(\vect{u}_{t})),\cdot)}{\partial \vect{g}}  \frac{\partial \vect{g}}{\partial \Psi}$

\State  $\nicefrac{\partial \mathcal{J}_{online}}{\partial\vect{\theta}} = \vect{v} + \vect{w}\vect{A}_{l,p}(\vect{u}_{t})$

\end{algorithmic}
\end{algorithm}

\subsection{Modification of the backward call}
The gradient of the online cost function can be computed by a modification of the backward call of modern automatic differentiation languages. The idea is to construct a ResNet-like computational graph using the sub-model $\vect{M}_{\vect{\theta}}$. We modify the gradient of the output of each ResNet block using a hook to include the information of the Jacobian of the non-differentiable solver. And the gradient of each block will correspond to $\nicefrac{\partial \vect{M}_{\vect{\theta}}(\cdot)}{ \partial \vect{\theta}}$. We provide an implementation of this technique, inspired by the syntax of PyTorch, in Algorithm \ref{algo2}.

\begin{algorithm}[h]
\caption{Gradient computation based on backward call modification}
\label{algo2}
\begin{algorithmic}
\Require
    \Statex $\Psi$: Non-differentiable solver of the hybrid system \eqref{eq:flow_discretized_equation}
    \Statex $\vect{M}_{\vect{\theta}}$: Deep learning based sub-model
    \Statex $\vect{u}_t$: Initial condition
    \Statex $n, h$: Number of simulation steps and time step
    \Statex $\mathcal{J}_{online}$ : Online cost function
    \Statex $l$ : Approximation scheme for the Jacobian of the flow

\Ensure   

\LineComment{\textit{Backward hook function}}    
    \Function{hook}{$\vect{z}_{t+jh}, j$}
    \LineComment{$\text{grad}(\cdot)$ \textit{refers to a modification of the gradient}}
    \State $\text{grad}(\vect{z}_{t+jh})$ = $h \cdot \text{grad}(\vect{z}_{t+nh}) \cdot \vect{J}_{j,l}$ \Comment\textit{{notice that $\vect{w} = \text{grad}(\vect{z}_{t+nh})$}}
    
    \EndFunction

\LineComment{\textit{Iterate through the solver $\Psi$}}
\For{$j \gets 1$ \textbf{to} $n$}

    \State $\vect{u}_{t + jh} = \Psi^j(\vect{u}_{t})$
\EndFor

\LineComment{\textit{Precompute the Jacobians $\vect{J}_{j,l}$}}
\For{$j \gets 1$ \textbf{to} $n-1$}

    \State $\vect{J}_{j,l} = \prod_{i = 1}^{i = n-j} \nicefrac{\partial \Psi(\Psi^{n-i}(\vect{u}_{t}))}{\partial \Psi^{n-i}(\vect{u}_{t})}$
\EndFor

\LineComment{\textit{Generate a ResNet like computational graph}}

\State $\vect{z}_{t} = \vect{u}_{t}$ \Comment{\textit{Initialize the ResNet state}}

\For{$j \gets 1$ \textbf{to} $n$}
    \State $\vect{z}_{t+jh} = \vect{M}_{\vect{\theta}}(\vect{z}_{t+(j-1)h})$ 
    \State $\text{data}(\vect{z}_{t+jh}) = \vect{u}_{t+jh}$ \Comment{$\text{data}(\cdot)$ \textit{refers to a modification of the value}}
    \If{$j\neq n$}
    \LineComment{\textit{Modify the gradient of the ResNet State}}
    \State  $\text{HOOK}(\vect{z}_{t+jh}, j)$ 
    \EndIf
\EndFor

\LineComment{\textit{Compute the online objective function}}

\State $Q(\cdot,\vect{g}(\Psi^n(\vect{u}_{t})),\vect{\theta}) = Q(\cdot,\vect{g}(\vect{z}(\vect{u}_{t+nh})),\vect{\theta})$ \Comment{\textit{The simulated states now have a computational graph}}

\State $\text{Backward}(Q(\cdot,\vect{g}(\Psi^n(\vect{u}_{t})),\vect{\theta}))$ \Comment{\textit{Run a backward call}}

\end{algorithmic}
\end{algorithm}

\section{Training configuration in the Lorenz 63 experiment}
\subsection{Training data}
In the first experiment with the Lorenz 63 system \ref{exp:order_conv}, we use multiple datasets $\mathcal{D}_{h} = \{(u^{\dagger}_{t_k +j{h}}, u^{\dagger}_{t_k})| \text{ with } k = 1 \dots N \text{ and } j = 1 \dots n\}$ that are sampled as the same time step as the Euler approximation. These datasets are generated using the LOSDA ODE solver \cite{odepack}. The number of training samples $N$ is equal to $100$ time steps and the number of simulation time steps $n$ is fixed to $10$. 

The dataset used in the second Lorenz 63 experiment \ref{exp:L63_pred}, $\mathcal{D}_{h} = \{(u^{\dagger}_{t_k +j{h_i}}, u^{\dagger}_{t_k})| \text{ with } k = 1 \dots N \text{ and } j = 1 \dots n\}$ is sampled at $h = 0.01$. The same sampling rate was used in multiple works \cite{brunton_discovering_2016,lguensat_analog_2017,ouala2020learning} that involve the data-driven identification of the Lorenz 63 system. This dataset is also generated using the LOSDA ODE solver \cite{odepack}. The number of training samples $N$ is equal to $5000$ time steps and the number of simulation time steps $n$ is fixed to $10$. 

\subsection{Training criterion and numerical solver}
In both the Lorenz 63 experiments, the online objective function corresponds to the mean squared error between the true Lorenz 63 state and the numerical integration of the model \eqref{eq:lorenz-63_approx} over $n = 10$ time steps. The cost function can be written as:
\begin{equation}
\label{eq:online_learning_L6396}
\mathcal{J} =  \frac{1}{N}\sum_k \frac{1}{n}\sum_{j = 1}^{n} \|{u}^{\dagger}_{t_{k} + jh} - \Psi^j({u}^{\dagger}_{t_k}) \|_2^2
\end{equation}
where $ \| \cdot \|_2$ is the L2 norm. The solver $\Psi$ used in these experiments is a differentiable DOPRI8 solver, developed in \cite{chen2018neural}.

\subsection{Parameterization of the deep learning sub-model and Baseline}
The sub-model employed in the Lorenz 63 experiments consists of a fully connected neural network with two hidden layers, each with three neurons. The activation function utilized for these hidden layers is the hyperbolic tangent. Below is a detailed description of the models tested in the experiment \ref{exp:L63_pred}:

\begin{itemize}
\item \textbf{No calibration and only using the physical core}: In this experiment we only run the physical core given by $\Dot{\vect{u}}_t = \vect{F}(\vect{u}_t)$ {\em i.e.}, by removing the sub-model $\vect{M}_{\vect{\theta}}$ in \eqref{eq:lorenz-63_approx}.

\item \textbf{Online calibration with exact gradient}: The sub-model is trained by utilizing the exact gradient of the online cost \eqref{eq:online_learning_L6396}. To compute the gradient of the solver, we rely on the fact that the solver used in this experiment is implemented in a differentiable language, allowing for automatic differentiation.

\item \textbf{Online calibration with a static approximation}: The sub-model is trained by utilizing the proposed Euler formulation of the gradient of the online cost \eqref{eq:online_learning_L6396}, in which we use a static approximation of the Jacobian as described in \eqref{eq:gradpsi_identity_corollary1}.

\item \textbf{Online calibration with an Ensemble approximation}: Similarly to above, the sub-model is trained by utilizing the proposed Euler formulation of the gradient of the online cost \eqref{eq:online_learning_L6396}, but the Jacobian is approximated using an ensemble as described in \ref{eq:EnJacobian}. The size of the ensemble is set to 5 members.
\end{itemize}

\subsection{Evaluation criteria}
In Figure \ref{fig:order_conv_L63}, the gradient error is computed as the mean absolute error between the exact gradient (computed using automatic differentiation) and the one returned by one of the proposed approximations. If we use equation \eqref{eq:grad_online_autodiff} to express this error and assuming that the exact gradient is $\frac{\partial \mathcal{J}}{\partial\vect{\theta}}$, the error depicted in Figure \ref{fig:order_conv_L63} is computed as:
\begin{equation}
\label{eq:grad_error}
\begin{aligned}
     \epsilon = \frac{1}{a} \| \frac{\partial \mathcal{J}}{\partial\vect{\theta}} - \vect{v} + \vect{w}\vect{A}_{l,p}(\vect{u}_{t}) \|_1
\end{aligned}    
\end{equation}
where $ \| \cdot \|_1$ is the L1 norm and $a$ is the number of parameters (we recall that $\vect{\theta} \in \mathbb{R}^a$). In this experiment, the tested gradients correspond to $p = 2$ (since we use a fixed number of simulation steps $n = 10$), and to $l = 1,2$ (which corresponds to the EGA and Static-EGA formulas in \eqref{eq:EGA_cases}). The exact Jacobian is in this experiment evaluated using automatic differentiation.

In Figure \ref{fig:L63_phase_space}, the MEG is simply the mean squared error at lead time $t_0+ih$ normalized by the initial error at $t_0+h$. It can be written as:
\begin{equation}
\label{eq:MEG_error}
\begin{aligned}
     \text{MEG}(t_0+ih)= \frac{\|  u^{\dagger}_{t_0 + ih} - \Psi^i(u^{\dagger}_{t_0})  \|_2^2}{\|  u^{\dagger}_{t_0 + h} - \Psi(u^{\dagger}_{t_0})  \|_2^2}
\end{aligned}    
\end{equation}

The Lyapunov spectrum in Table \ref{tab:fore_Lor_topo} is computed using the Gram-Schmidt orthonormalization technique \cite{Parker1989_stability}, and the Lyapunov dimension is deduced from the spectrum as given in \cite{Parker1989_dim}.

\section{Training configuration in the QG experiment}
\subsection{traning data}

\begin{table}
    \centering
    \begin{tabular}{c|c|c|c|c|c|c|c|c}
         \thead{DNS grid \\($N_{\text {DNS }} \times N_{\text {DNS }}$)  }& \thead{LES grid \\($N_{\text {LES }} \times N_{\text {LES }}$)}& \thead{Scale \\($\delta$) }& \thead{DNS time step \\$(h_{DNS})$ }&\thead{ LES time step  \\$(h_{LES})$} & $\operatorname{Re}$  & $r$ & $k_f$ \\
         \hline
         $1024 \times 1024$ & $64 \times 64$ & $16$ & $5 \times 10^{-5}$ & $8 \times 10^{-4}$ & $20000$ & $0.1$  &  $4$\\
    \end{tabular}
    \caption{{{{\bf \em Parameters of the DNS Flow Configurations}. Both the DNS and LES systems share the same parameters, except for the grid size and the integration time step. The grid size of the LES system is reduced by a factor of $\delta = 16$. The time step of the LES system corresponds to that of the DNS multiplied by $\delta$.}}}
    \label{tab:flow_configuration_QG}
\end{table}

We run the QG equations with the flow configuration given in Table \ref{tab:flow_configuration_QG} starting from 14 different initial random fields of vorticity on a high-resolution grid to generate direct numerical simulation (DNS) data. These runs are used to generate 14 different initial conditions that are in the statistical equilibrium regime. The new initial conditions are then used to generate DNS data of 2 million time steps that correspond to 4000 eddy turnover times. These data are then filtered to the resolution of the LES simulation and sampled every $h_{LES}$. From the 14 runs, we used 8 datasets for training one for validation, and the remaining 5 datasets for testing and evaluation of the models. Regarding the training and validation datasets, they are written as $\{(\overline{\omega}_{t_k +j{h_{LES}}}, (\overline{\omega}_{t_k},\overline{\psi}_{t_k}))| \text{ with } k = 1 \dots N \text{ and } j = 1 \dots n\}$ where $N = 2000 \times 8$ (where 8 is the number of training datasets) and $n = 10$ for the online learning experiments and as $\{(\Pi_{t_k}, (\overline{\omega}_{t_k},\overline{\psi}_{t_k}))| \text{ with } k = 1 \dots N\}$. 

\subsection{Training criterion}
The online objective function in the QG experiment corresponds to the means squared error between the true vorticity field and the one issued from the numerical integration of the model \eqref{eq:QG_LES} over $n = 10$ time steps. The cost function can be written as:
\begin{equation}
\label{eq:online_learning_QG}
\mathcal{J} =  \frac{1}{N}\sum_k \frac{1}{n}\sum_{j = 1}^{n} \|\overline{\omega}_{t_{k} + jh} - \Psi^j(\overline{\omega}_{t_k}) \|
\end{equation}
The solver $\Psi$ depends solely on the vorticity field $\overline{\omega}$ as all the other variables can be deduced from $\overline{\omega}$.

The offline objective function is the mean squared error between the output of the sub-model $\vect{M}_{\vect{\theta}}$ and the reference subgrid scale term $\Pi$. It can be written as:
\begin{equation}
\label{eq:offline_leqrning_QG}
\mathcal{J} =  \frac{1}{N}\sum_k \frac{1}{n}\sum_{j = 1}^{n} \|\Pi_{t_k} - \vect{M}_{\vect{\theta}}(\overline{\omega}_{t_k}, \overline{\psi}_{t_k}) \|
\end{equation}

\subsection{Numerical Solver of the QG system}
QG system is solved using a code adapted from \cite{frezat2022posteriori}. This solver is written in a differentiable language which allows the comparison of our proposed approximation to models that are optimized online with the true gradient of the solver. It relies on a pseudo-spectral solver and a classical fourth-order Runge-Kutta time integration scheme. 

\subsection{Filtering and coarse graining operation}
The QG equations are defined on a double periodic squared domain $\Omega \in[-\pi, \pi]^2$. The DNS solution is constructed on a regular $N_{\text {DNS }} \times N_{\text {DNS }}$ grid with a uniform spacing $\Delta_{\text {DNS }}=2\pi N_{\text {DNS }}^{-1}$. The LES system of equations is obtained by projecting the DNS states through a convolution with a spatial kernel $G$, followed by a discretization on the reduced grid, with larger spacing $\Delta_{\text {LES }}=\delta \Delta_{\text {DNS }}$. We use in this experiment a Gaussian filter that can be defined in spectral space as:
$$
G_{\nu}=\exp \left(-\frac{{\nu}^2 \Delta_{\text {f }}^2}{24}\right), \\
$$
where $\Delta_f$ is the filter size, which is taken to be $\Delta_f = 2 \Delta_{\text {LES }}$ to yield sufficient resolution \cite{guan2023learning}. This filtering/coarse-graining operation can be written as (taking here as an example the DNS vorticity field $\omega$):
$$
\bar{\omega}_{\nu}=\left(\omega * G\right)\left(|{\nu}|<\pi \Delta_{\text {LES }}^{-1}\right) .
$$
Regarding numerical aspects, we can solve the time integration of the LES system with a larger time-step by a factor corresponding to the grid size ratio, that is, $\Delta t_{\text {LES }}=\delta \Delta t_{\text {DNS }}$.

\section{Parameterization of the deep learning sub-model and baseline}
The sub-model used in the QG experiments is a Convolutional Neural Network (CNN) that has the same architecture as the one used in \cite{guan2022stable}. The inputs of the CNN are the vorticity and streamfunction fields $(\overline{\omega}_{t_k}, \overline{\psi}_{t_k})$. The convolutional layers have the same dimension $(64 \times 64)$ as that of the input and output layers. All layers are initialized randomly. The number of channels is set to 64 and the filter size is $(5 \times 5)$. The activation function of each layer is ReLu (rectified linear unit) except for the last one, which is a linear map. Similarly to \cite{frezat2022posteriori} We use periodic padding.

Below is a detailed description of the models tested in the QG experiment: 
\begin{itemize}

\item \textbf{Online calibration with exact gradient}: In this calibration scheme, we assume that the solver of \eqref{eq:QG_LES} is differentiable and we optimize the parameters of the sub-model with the exact gradient of the online objective function. This experiment was already studied in \cite{frezat2022posteriori} and shows better stability performance than offline calibration schemes.

\item \textbf{Online calibration with a static approximation}: In this experiment, we evaluate the performance of the proposed online learning scheme in which the gradient of the online loss function is approximated based on \eqref{eq:gradpsi_identity_corollary1}. In this experiment, the solver of \eqref{eq:QG_LES} is not assumed to be differentiable.

\item \textbf{Offline calibration}: We also compare the proposed static approximation to a simple offline learning strategy in which the CNN is calibrated to reproduce the subgrid-scale term $\Pi_t$.

\item \textbf{Physical SGS model with dynamic Smagorinsky (DSMAG)}: We also evaluate and compare the proposed approximation schemes with respect to classical physics-based parameterization given by the dynamic Smagorinsky (DSMAG) model. In this model, the impact of the unresolved scales on the dynamics is assumed to be diffusive with a diffusion constant that is computed automatically. The dynamic Smagorinsky model has been widely used as a baseline in many studies. For a detailed explanation, please refer to, for example, \cite{guan2022stable}. 
\end{itemize}

\subsection{Additional experiment on the two scale Lorenz 96}
The two scale L96 system describes a coupled system of equations \cite{Lorenz96_paper} with $S$ slow variables, $u^{\dagger}_{t} = [u^{\dagger}_{t,1}, u^{\dagger}_{t,2}, \cdots, u^{\dagger}_{t,S}]^T$ each of which is coupled to $B$ fast variables $(y_{t,1,s},y_{t,2,s}, \cdots, y_{t,B,s})$: 
\begin{equation}
\label{eq:True_2SL96}
\begin{aligned}
\Dot{{u}}^{\dagger}_{t,s}  &= - {{u}^{\dagger}}_{t,{s-1}} \left(  {{u}^{\dagger}}_{t,{s-2}} -  {{u}^{\dagger}}_{t,{s+1}} \right) - {{u}^{\dagger}}_{t,{s}} + A + R_{t,s}\\
\Dot{{y}}_{t,b,s} &= - c\gamma{y}_{t,{b+1,s}} \left( {y}_{t,{b+2,s}} - {y}_{t,{b-1,s}} \right) - c {y}_{t,{b,s}} + \frac{dc}{\gamma}  {{u}^{\dagger}}_{t,{s}}
\end{aligned}
\end{equation}
where $R_{t,s}= - \left( \frac{dc}{\gamma} \right) \sum_{b=1}^{B} {y}_{t,{b,s}}$

In this experiment, We assume that the physical core $\vect{F}$ represents the equations that govern the slow variables and we use a sub-model $\vect{M}_{\vect{\theta}}$ to mimic the impact of the fast variables $y_{t,b,s}$:
\begin{equation}
\label{eq:lorenz-96_approx}
\Dot{\vect{u}}_t = \vect{F}(\vect{u}_t) + \vect{M}_{\vect{\theta}}(\vect{u}_t)
\end{equation}
where $\vect{u}_t = [{u}_{t,1}, {u}_{t,1}, \cdots, {u}_{t,S}]^T \in \mathbb{R}^S$ and $\vect{M}_{\vect{\theta}}$ is a fully connected neural network with parameters $\theta$. The physical core $\vect{F} = [{F}_1, {F}_2, \cdots, {F}_S]^T$ is given by:
\begin{equation}
\label{eq:Slow_L96}
\Dot{{u}}_{t,s}  = {F}_s = -  {{u}^{\dagger}}_{t,{s-1}} \left(  {{u}^{\dagger}}_{t,{s-2}} -  {{u}^{\dagger}}_{t,{s+1}} \right) - {{u}^{\dagger}}_{t,{s}} + A
\end{equation}

\begin{figure*}
\centering
\includegraphics[width=1\textwidth]{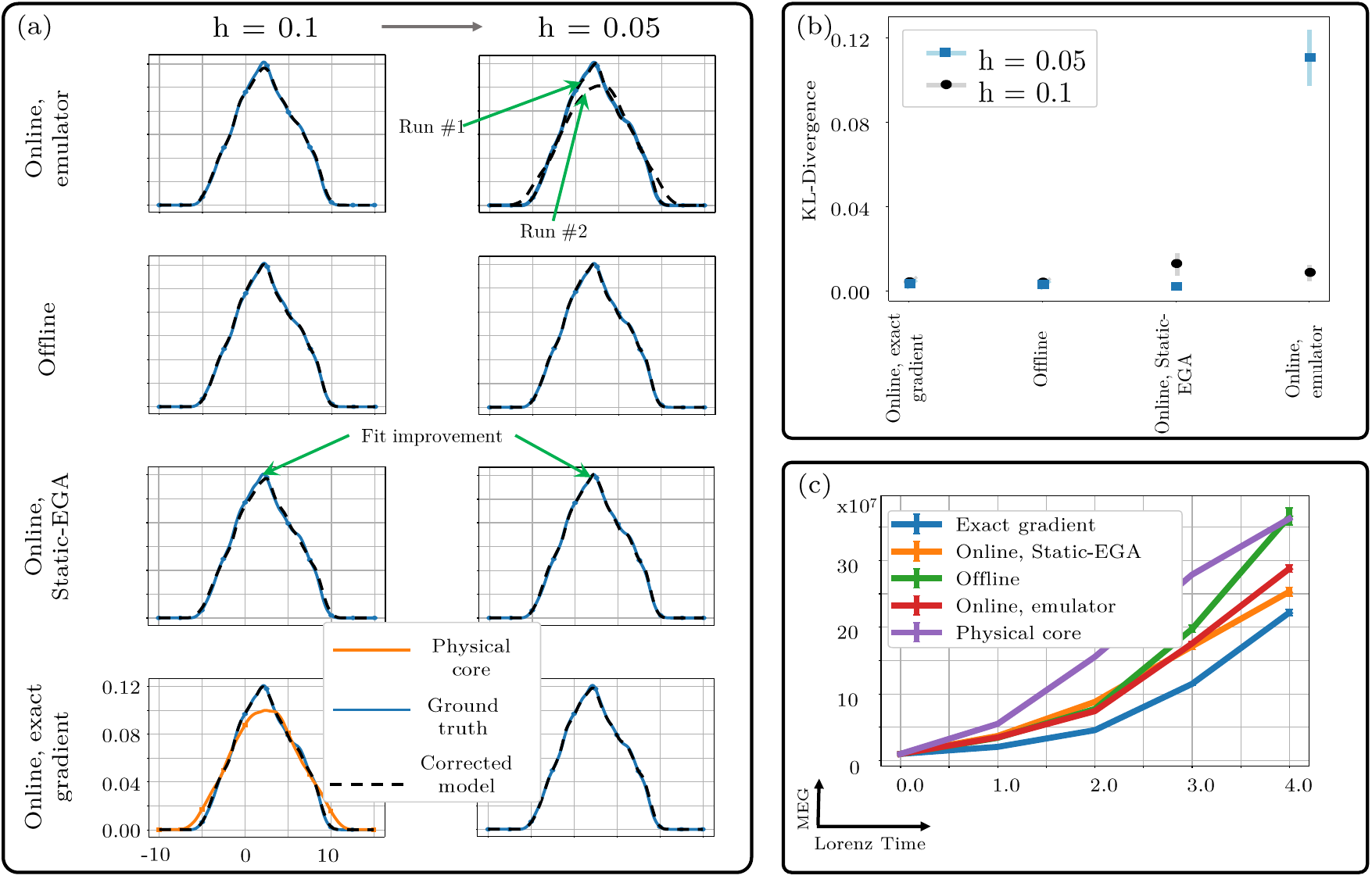}
\caption{{{{ \bf  \em Simulation example of the tested models in the Lorenz 96 esperiment}. We compare a multiscale simulation of the true system \eqref{eq:True_2SL96} to the physical core in \eqref{eq:lorenz-63_approx} (without the neural network sub-model) and to corrected models where the correction term $\vect{M}_{\vect{\theta}}$ is calibrated both online and offline. We compare the PDF of the first state of the tested models with respect to the one computed from a simulation of the true system given by \eqref{eq:True_2SL96} both qualitatively in (a) and quantitatively in (b). (c) Mean error growth of the tested models. We plot both the mean and standard deviation that was computed based on an ensemble of 20 trajectories issued from 5 different learning realizations. The error bars are scaled by 1/20.}
}}
\label{fig:L96_experiment}
\end{figure*}

The goal of this experiment is to evaluate the proposed Euler approximation of the online learning of the sub-model $\vect{M}_{\vect{\theta}}$ (based on a static approximation of the Jacobian of the flow) on a multiscale dynamical systems, for varying time steps of the Euler approximation. We set the number of slow variables $S = 8$ and the number of fast variables $B = 5$. Regarding the values of the parameters of the equation, we use the following configuration: $A = 8$, $d = 1$, $\gamma = 10$, and $c = 10$. This simulation configuration yields chaotic dynamics, where the statistical properties can not be solely explained by the slow model \eqref{eq:Slow_L96} (this can be visualized in Figure \ref{fig:L96_experiment}, where the PDF of the physical core is given with respect to the one of the multiscale system). We assume here that the number of time steps $n$ is fixed, and we run a series of two experiments for which the time step of the Euler approximation of the gradients decreases from $h = 0.1$ to $h = 0.05$. 

We compare the proposed Euler approximation to both offline and online calibration techniques. The online calibration is carried using an exact gradient and also with an emulator. The emulator is trained sequentially to the sub-model $\vect{M}_{\vect{\theta}}$ as discussed in section \eqref{sec:emulators}. In the online learning experiments, the training datasets correspond to time series of the full Lorenz 96 system $\{(u^{\dagger}_{t_k +j{h_i}}, u^{\dagger}_{t_k})| \text{ with } k = 1 \dots N \text{ and } j = 1 \dots n\}$ and in the offline learning experiment, the training data corresponds to $\{({u}^{\dagger}_{t_k},  \vect{R}_{t_k} = [{R}_{t_k,1}, {R}_{t_k,2}, \cdots, {R}_{t_k,S})]^T| \text{ with } k = 1 \dots N \}$. The number of simulation steps is set to $n = 10$ and the size of the dataset $N$ is equal to $20000$.

Overall, the following models are tested:
\begin{itemize}

\item \textbf{Online calibration with exact gradient}: In this calibration scheme, we implement the solver of \eqref{eq:lorenz-96_approx} in a differentiable language and we optimize the parameters of the sub-model with the exact gradient of the online objective function. 

\item \textbf{Online calibration with a static approximation}: In this experiment, we evaluate the performance of the proposed online learning scheme in which the gradient of the online loss function is approximated based on the Static-EGA \eqref{eq:gradpsi_identity_corollary1}. In this experiment, the solver of \eqref{eq:lorenz-96_approx} is not assumed to be differentiable.

\item \textbf{Online calibration with an emulator}: We also evaluate and compare the proposed approximation schemes to online learning with emulators as presented in section \ref{sec:emulators}. We recall that in this experiment, physical core $\vect{F}$ in \eqref{eq:lorenz-96_approx} is replaced (in the training phase) by a neural network that is trained sequentially with $\vect{M}_{\vect{\theta}}$. This network is a linear quadratic model, similar to the one discussed in \cite{fablet_blin_ieee}.

\item \textbf{Offline calibration}: We also compare the proposed static approximation to a simple offline learning strategy. In this experiment, the parameters of the sub-model are optimized to minimize the following offline objective function:
\begin{equation}
\label{eq:offline_learning_L96}
   \mathcal{Q} = \frac{1}{N}\sum_k \left \|  \vect{R}_{t_k}  -  \vect{M}_{\vect{\theta}}({{u}^{\dagger}_{t_k}})   \right \|^2
\end{equation}

\end{itemize}
Besides the offline calibration scheme, the online cost function used in this experiment is the mean squared error of the numerical integration of \eqref{eq:lorenz-96_approx} with respect to the true two-scale Lorenz 96 sequence. All the tested experiments share the same parameterization of the sub-model $\vect{M}_{\vect{\theta}}$, which is a fully connected neural network with 6 hidden layers, each with 100 neurons and a hyperbolic tangent activation.

\subsubsection{Performance of the learnt sub-model}
We plot the performance of both online and offline optimization approaches in the Lorenz 96 case study in Figure \ref{fig:L96_experiment}. We evaluate the approaches with respect to the true Lorenz 96 simulation and also with respect to a simulation issued from the physical core. Overall, we notice that both offline and online schemes noticeably outperform the physical core which highlights the relevance of using such corrections. Regarding the short-term prediction performance, panel (c) of Figure \ref{fig:L96_experiment} shows that all models are able to provide better predictions than the physical core. in this experiment, the online learning with true gradient is able to provide the best prediction performance. When evaluating the qualitative properties of the simulation of the models highlighted for instance in panels (a) and (b) in Figure \ref{fig:L96_experiment}, We also found that the proposed approximate gradient for the online learning scheme provides a very nice correction that is on the same level as online optimization with the true gradient. The proposed approximation also improves when we reduce the time step of the Euler approximation of the gradient from $h = 0.1$ to $h = 0.05$. This experiment also reveals that online learning with an emulator can be challenging. Specifically, the qualitative and quantitative comparison of the PDF of the models trained online with an emulator in Figure \ref{fig:L96_experiment} shows that the sub-model calibration can be highly sensitive and shows that two different model initializations can lead to very distinct sub-models. 

\section{Proof of the equation \eqref{eq:euler_grad_psitheorem_psi}}
\label{seq:prof_recu}
The proof of \eqref{eq:euler_grad_psitheorem_psi} is conducted by recurrence. For $n = 1$ we have:
\begin{equation}
\label{eq:rec_proof_n1}
\begin{aligned}
\frac{\partial}{\partial\vect{\theta}}  {\Psi^1}(\vect{u}_{t}) &= {\frac{\partial \Psi({\Psi^{0}}(\vect{u}_t))}{\partial\vect{\theta}}}
\end{aligned}
\end{equation}
which is by inspection of \eqref{eq:gradient_of_psi_elementaire_noO} true.

Assume that \eqref{eq:euler_grad_psitheorem_psi} is true for all $n$, for $n+1$ we have:
\begin{equation}
\label{eq:euler_grad_psi}
\begin{aligned}
\frac{\partial}{\partial\vect{\theta}}  {\Psi^{n+1}}(\vect{u}_{t}) &= \sum_{j = 1}^{j = n} (\prod_{i = 1}^{i = n+1-j}  \frac{\partial \Psi(\Psi^{n+1-i}(\vect{u}_{t}))}{\partial \Psi^{n+1-i}(\vect{u}_{t})})\frac{\partial}{\partial\vect{\theta}} \Psi(\Psi^{j -1}(\vect{u}_{t})) + \frac{\partial}{\partial\vect{\theta}} \Psi(\Psi^{n}(\vect{u}_{t}))\\
&= \prod_{i = 1}^{i = n} \frac{\partial \Psi(\Psi^{n+1-i}(\vect{u}_{t}))}{\partial \Psi^{n+1-i}(\vect{u}_{t})} \frac{\partial}{\partial\vect{\theta}} \Psi(\Psi^{0}(\vect{u}_{t}))\\
&+ \prod_{i = 1}^{i = n-1} \frac{\partial \Psi(\Psi^{n+1-i}(\vect{u}_{t}))}{\partial \Psi^{n+1-i}(\vect{u}_{t})} \frac{\partial}{\partial\vect{\theta}} \Psi(\Psi^{1}(\vect{u}_{t}))\\
&\vdots\\
&+ \prod_{i = 1}^{i = 2} \frac{\partial \Psi(\Psi^{n+1-i}(\vect{u}_{t}))}{\partial \Psi^{n+1-i}(\vect{u}_{t})} \frac{\partial}{\partial\vect{\theta}} \Psi(\Psi^{n-2}(\vect{u}_{t}))\\
&+ \prod_{i = 1}^{i = 1} \frac{\partial \Psi(\Psi^{n+1-i}(\vect{u}_{t}))}{\partial \Psi^{n+1-i}(\vect{u}_{t})} \frac{\partial}{\partial\vect{\theta}} \Psi(\Psi^{n-1}(\vect{u}_{t}))\\
&+ \frac{\partial}{\partial\vect{\theta}} \Psi(\Psi^{n}(\vect{u}_{t}))\\
\end{aligned}
\end{equation}
if we take $\frac{\partial \Psi(\Psi^{n}(\vect{u}_{t}))}{\partial \Psi^{n}(\vect{u}_{t})} $ as a common factor we have: 
\begin{equation}
\begin{aligned}
\frac{\partial}{\partial\vect{\theta}}  {\Psi^{n+1}}(\vect{u}_{t}) &= \frac{\partial \Psi(\Psi^{n}(\vect{u}_{t}))}{\partial \Psi^{n}(\vect{u}_{t})}\prod_{i = 2}^{i = n} \frac{\partial \Psi(\Psi^{n+1-i}(\vect{u}_{t}))}{\partial \Psi^{n+1-i}(\vect{u}_{t})} \frac{\partial}{\partial\vect{\theta}} \Psi(\Psi^{0}(\vect{u}_{t}))\\
&+ \frac{\partial \Psi(\Psi^{n}(\vect{u}_{t}))}{\partial \Psi^{n}(\vect{u}_{t})}\prod_{i =2}^{i = n-1} \frac{\partial \Psi(\Psi^{n+1-i}(\vect{u}_{t}))}{\partial \Psi^{n+1-i}(\vect{u}_{t})} \frac{\partial}{\partial\vect{\theta}} \Psi(\Psi^{1}(\vect{u}_{t}))  \\
&\vdots\\
&+ \frac{\partial \Psi(\Psi^{n}(\vect{u}_{t}))}{\partial \Psi^{n}(\vect{u}_{t})}\prod_{i = 2}^{i = 2} \frac{\partial \Psi(\Psi^{n+1-i}(\vect{u}_{t}))}{\partial \Psi^{n+1-i}(\vect{u}_{t})}  \frac{\partial}{\partial\vect{\theta}} \Psi(\Psi^{n-2}(\vect{u}_{t}))  \\
&+ \frac{\partial \Psi(\Psi^{n}(\vect{u}_{t}))}{\partial \Psi^{n}(\vect{u}_{t})}  \frac{\partial}{\partial\vect{\theta}}\Psi(\Psi^{n-1}(\vect{u}_{t})) \\
&+ \frac{\partial}{\partial\vect{\theta}} \Psi(\Psi^{n}(\vect{u}_{t}))\\
\end{aligned}
\end{equation}
Factorization of $\frac{\partial \Psi(\Psi^{n}(\vect{u}_{t}))}{\partial \Psi^{n}(\vect{u}_{t})}$:
\begin{equation}
\begin{aligned}
\frac{\partial}{\partial\vect{\theta}}  {\Psi^{n+1}}(\vect{u}_{t}) = \frac{\partial \Psi(\Psi^{n}(\vect{u}_{t}))}{\partial \Psi^{n}(\vect{u}_{t})}( & \prod_{i = 2}^{i = n} \frac{\partial \Psi(\Psi^{n+1-i}(\vect{u}_{t}))}{\partial \Psi^{n+1-i}(\vect{u}_{t})}  \frac{\partial}{\partial\vect{\theta}} \Psi(\Psi^{0}(\vect{u}_{t})) \\
+ &\prod_{i = 2}^{i = n-1} \frac{\partial \Psi(\Psi^{n+1-i}(\vect{u}_{t}))}{\partial \Psi^{n+1-i}(\vect{u}_{t})} \frac{\partial}{\partial\vect{\theta}} \Psi(\Psi^{1}(\vect{u}_{t})) \\
&\vdots\\
+ &\prod_{i = 2}^{i = 2} \frac{\partial \Psi(\Psi^{n+1-i}(\vect{u}_{t}))}{\partial \Psi^{n+1-i}(\vect{u}_{t})} \frac{\partial}{\partial\vect{\theta}} \Psi(\Psi^{n-2}(\vect{u}_{t}))\\
&+ \frac{\partial}{\partial\vect{\theta}} \Psi(\Psi^{n-1}(\vect{u}_{t})) )\\
&+ \frac{\partial}{\partial\vect{\theta}} \Psi(\Psi^{n}(\vect{u}_{t})) \\
\end{aligned}
\end{equation}

Which is conveniently written as:
\begin{equation}
\label{eq:eqforproof}
\begin{aligned}
\frac{\partial}{\partial\vect{\theta}}  {\Psi^{n+1}}(\vect{u}_{t}) = \frac{\partial \Psi(\Psi^{n}(\vect{u}_{t}))}{\partial \Psi^{n}(\vect{u}_{t})}( & \prod_{i = 1}^{i = n-1} \frac{\partial \Psi(\Psi^{n-i}(\vect{u}_{t}))}{\partial \Psi^{n-i}(\vect{u}_{t})}  \frac{\partial}{\partial\vect{\theta}} \Psi(\Psi^{0}(\vect{u}_{t})) \\
+ &\prod_{i = 1}^{i = n-2} \frac{\partial \Psi(\Psi^{n-i}(\vect{u}_{t}))}{\partial \Psi^{n-i}(\vect{u}_{t})} \frac{\partial}{\partial\vect{\theta}} \Psi(\Psi^{1}(\vect{u}_{t})) \\
&\vdots\\
+ &\prod_{i = 1}^{i = 1} \frac{\partial \Psi(\Psi^{n-i}(\vect{u}_{t}))}{\partial \Psi^{n-i}(\vect{u}_{t})} \frac{\partial}{\partial\vect{\theta}} \Psi(\Psi^{n-2}(\vect{u}_{t}))\\
&+ \frac{\partial}{\partial\vect{\theta}} \Psi(\Psi^{n-1}(\vect{u}_{t})) )\\
&+ \frac{\partial}{\partial\vect{\theta}} \Psi(\Psi^{n}(\vect{u}_{t})) \\
\end{aligned}
\end{equation}
Equation \eqref{eq:eqforproof} can be written in a more compact form as:
\begin{equation}
\begin{aligned}
\frac{\partial}{\partial\vect{\theta}}  {\Psi^{n+1}}(\vect{u}_{t}) &= \frac{\partial \Psi(\Psi^{n}(\vect{u}_{t}))}{\partial \Psi^{n}(\vect{u}_{t})}\underbrace{( \sum_{j = 1}^{j = n-1} (\prod_{i = 1}^{i = n-j} \frac{\partial \Psi(\Psi^{n-i}(\vect{u}_{t}))}{\partial \Psi^{n-i}(\vect{u}_{t})})  \frac{\partial}{\partial\vect{\theta}} \Psi(\Psi^{j -1}(\vect{u}_{t})) + \frac{\partial}{\partial\vect{\theta}} \Psi(\Psi^{n-1}(\vect{u}_{t})) )}_{= \frac{\partial}{\partial\vect{\theta}}  {\Psi^n}(\vect{u}_{t})}\\
&+ \frac{\partial}{\partial\vect{\theta}} \Psi(\Psi^{n}(\vect{u}_{t}))\\
&= \frac{\partial \Psi(\Psi^{n} (\vect{u}_{t}))}{\partial \Psi^{n} (\vect{u}_{t})} \frac{\partial \Psi^{n} (\vect{u}_{t})}{\partial\vect{\theta}} +  {\frac{\partial \Psi(\Psi^{n} (\vect{u}_{t}))}{\partial\vect{\theta}}}
\end{aligned}
\end{equation}
which is by inspection of \eqref{eq:gradient_of_psi_elementaire_noO} true. This completes the proof of the formula \eqref{eq:euler_grad_psitheorem_psi}.

\section{Data and code availability}
The code developed for this work is written in Python using Pytorch and is available at \url{https://github.com/CIA-Oceanix/EGA}.

\end{document}